%% file: main.tex
\pdfoutput=1

\documentclass[11pt]{article}

\usepackage[final]{acl}

\usepackage{times}
\usepackage{latexsym}
\usepackage{booktabs}
\usepackage{pifont}
\usepackage{subcaption} 
\usepackage{svg}
\usepackage{amsmath}
\usepackage{mathtools}
\usepackage{amssymb}
\usepackage{cleveref} 
\usepackage{algorithm}
\usepackage{mathrsfs}
\usepackage{xcolor}
\usepackage{adjustbox}
\usepackage{wrapfig}
\usepackage{algpseudocode}
\usepackage{tabularx}
\usepackage{mathbbol}
\usepackage{tcolorbox}
\usepackage{caption}
\usepackage{float}
\usepackage{lipsum}
\usepackage{xtab}
\usepackage{makecell}
\usepackage{xcolor}
\usepackage{verbatim}
\usepackage{lipsum} 
\usepackage{multicol}
\definecolor{lowcontrastgreen}{RGB}{160,206,86}
\definecolor{lowcontrastred}{RGB}{255,95,83}
\definecolor{lowcontrastyellow}{RGB}{255,218,93}
\definecolor{lowcontrastblue}{RGB}{181,234,215}
\usepackage[T1]{fontenc}
\newcommand{\cmark}{\ding{51}}  
\newcommand{\xmark}{\ding{55}}  
\usepackage[utf8]{inputenc}
\usepackage{boldline}
\usepackage{microtype}
\usepackage{multicol}
\usepackage{multirow}
\usepackage{inconsolata}
\usepackage{svg}
\usepackage{graphicx}
\usepackage{listings}
\usepackage{epsfig,endnotes}
\usepackage[utf8]{inputenc}
\usepackage[T1]{fontenc}
\usepackage{upquote}

\title{Beyond Perception: Evaluating Abstract Visual Reasoning through Multi-Stage Task}

\author{
 \textbf{Yanbei Jiang\textsuperscript{1}},
 \textbf{Yihao Ding\textsuperscript{1,2}},
 \textbf{Chao Lei\textsuperscript{1}},
 \textbf{Jiayang Ao\textsuperscript{1}}\\
 \textbf{Jey Han Lau\textsuperscript{1}},
 \textbf{Krista A. Ehinger\textsuperscript{1}}
\\
\textsuperscript{1}The University of Melbourne,
\textsuperscript{2}University of Sydney
\\
 \small{
\href{mailto:email@domain}{yanbeij@student.unimelb.edu.au}} \\
\small{
\href{mailto:email@domain}{jeyhan.lau@gmail.com}} \\
\small{
\href{mailto:email@domain}{
kehinger@unimelb.edu.au}
 }
}

\begin{document}
\maketitle
\begin{abstract}
Current Multimodal Large Language Models (MLLMs) excel in general visual reasoning but remain underexplored in Abstract Visual Reasoning (AVR), which demands higher-order reasoning to identify abstract rules beyond simple perception. Existing AVR benchmarks focus on single-step reasoning, emphasizing the end result but neglecting the multi-stage nature of reasoning process. Past studies found MLLMs struggle with these benchmarks, but it doesn't explain how they fail. To address this gap, we introduce MultiStAR, a \textbf{M}ulti-\textbf{St}age \textbf{A}V\textbf{R} benchmark based on RAVEN, to assess reasoning across varying levels of complexity. Additionally, existing metrics like accuracy only focus on the final outcomes while do not account for the correctness of intermediate steps. Therefore, we propose a novel metric, MSEval, which considers the correctness of intermediate steps in addition to the final outcomes. We conduct comprehensive experiments on MultiStAR using 17 representative close-source and open-source MLLMs. The results reveal that while existing MLLMs perform adequately on basic perception tasks, they continue to face challenges in more complex rule detection stages. The dataset and code are available at \href{https://github.com/YanbeiJiang/MultiStAR}{https://github.com/YanbeiJiang/MultiStAR}
\end{abstract}

\begin{figure*}[ht]
    \centering
    \includegraphics[width=0.8\textwidth]{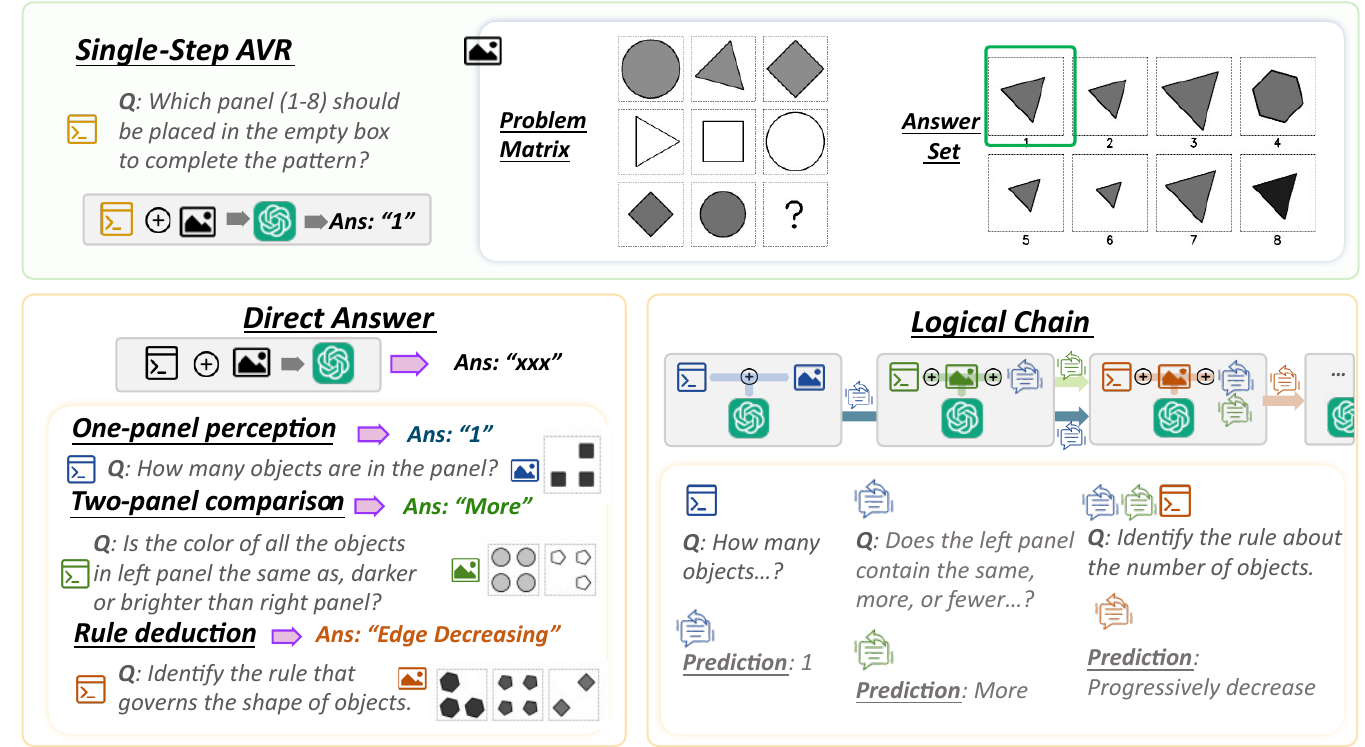}
    \caption{{Left}: RAVEN puzzle. The correct answer is 1. {Right}: Direct Answer subtask, where questions are independent for each configuration; Logical Chain subtask, where information from previous stages is used to assist in answering the current stage. All questions here focus on the concept of Number.}
    \label{fig:intro_image}
\end{figure*}

\section{Introduction}
Multimodal Large Language Models (MLLMs) demonstrate proficiency in addressing a wide array of visual-text inquiries and show strong multimodal understanding ability in tasks such as visual question answering \cite{goyal2017making,marino2019ok,ding2023vqa}, image captioning \cite{saito2023pic2word,vinyals2015show,jiang2024kale}, and visual grounding \cite{he2024improved,deng2021transvg}. These tasks focus on evaluating the models' capability to understand real-world or domain-specific knowledge. However, Abstract Visual Reasoning (AVR) presents a different challenge, focusing on a model’s ability to identify and reason through abstract patterns, relationships, and rules. A well-known example of AVR tasks is RAVEN \cite{raven2003raven, zhang2019raven}, as shown in the left half of Figure \ref{fig:intro_image}. 
The solver needs to select the correct panel from a answer set to complete a 3x3 problem matrix by deducing the visual rules governing the grid's arrangement. For instance, by analyzing the colors of each panel, one might observe the color remains consistent across each row. Unlike other multimodal tasks in real-world scenarios, AVR focuses on reasoning about arbitrary visual elements, serving as a robust benchmark for evaluating the zero-shot reasoning capabilities of MLLMs in visual contexts \cite{mandziuk2019deepiq,santoro2018measuring}. 

Previous works have consistently shown that AVR tasks pose challenges for MLLMs in zero-shot inference settings. Despite recent advancements like Chain-of-Thought prompting \cite{ahrabian2024curious,gendron2024large} and the inclusion of oracle captions \cite{zhang2024far}, models continue to perform at near-random levels on these tasks. The AVR datasets used commonly in these evaluations like RAVEN primarily focus on single-step end-to-end reasoning (i.e., giving the models the questions and asking them to derive the final answer; left half of Figure \ref{fig:intro_image}) \cite{santoro2018measuring,nie2020bongard,cao2024visual}. However, this design deviates from the human reasoning process, which often involves sequential steps: starting with single-panel perception, progressing to panel comparisons, and finally deducing the underlying rules before solving the puzzle. Previous datasets often omit these intermediate stages, and as such cannot evaluate MLLM's step-by-step reasoning capabilities and identify where they struggle within the reasoning process. This highlights the need for benchmarks that \textit{assess intermediate perception and reasoning processes}. Additionally,
existing evaluation metrics like accuracy measure only the performance of the final answer without \textit{considering the correctness of intermediate steps.}, and so they are unable to assess partial success.

To address these limitations, we introduce MultiStAR, a Multi-Stage Abstract Visual Reasoning dataset based on RAVEN, to evaluate MLLMs on the intermediate steps in the reasoning process. As shown in the right half of Figure \ref{fig:intro_image}, the dataset is divided into two sub-tasks, each focusing on different aspects of reasoning. The first sub-task, referred to as \textbf{Direct Answer}, evaluates model performance at varying levels of complexity to assess perception and reasoning abilities at each individual step. Using template-based methods, we generate questions based on RAVEN, ranging from basic object recognition to advanced comparison, pattern recognition, and rule inference. This approach ensures comprehensive coverage of reasoning patterns. The second sub-task called \textbf{Logical Chain}, emphasizes how models measure and maintain logical correlations across reasoning steps. Using puzzles from the RAVEN as the final question, we decompose the reasoning process into a sequence of subproblems in a bottom-up manner. Each stage in this chain links the current reasoning task to its dependent subproblems, requiring the model to combine current information with outputs from previous stages. 
To assess the correctness of intermediate steps, we introduce a novel metric, MSEval, which provides a more fine-grained assessment of the model's reasoning process for the logical chain task. MSEval uses the correct answer probabilities at each stage to compute the joint probability across the reasoning process. This approach considers both the correctness of the current stage and all dependent intermediate steps.
\begin{table*}[ht]
\centering
\resizebox{\textwidth}{!}{
\begin{tabular}{lccccccc}
\toprule
Dataset    & \multicolumn{1}{c}{\begin{tabular}[c]{@{}c@{}}Num. of \\ Images\end{tabular}} & \multicolumn{1}{c}{\begin{tabular}[c]{@{}c@{}}Num. of \\ QA pairs\end{tabular}} & 
\multicolumn{1}{c}{Reasoning Domain (Task Focus)} & 
\multicolumn{1}{c}{\begin{tabular}[c]{@{}c@{}}Question \\ Generation\end{tabular}} & \multicolumn{1}{c}{\begin{tabular}[c]{@{}c@{}}Answer \\ Type\end{tabular}} & \multicolumn{1}{c}{\begin{tabular}[c]{@{}c@{}}Functional \\ Program\end{tabular}} & \multicolumn{1}{c}{\begin{tabular}[c]{@{}c@{}}Multi-Step \\ Structure\end{tabular}} \\ 
\midrule
CLEVR \cite{johnson2017clevr}     & 100K     & 1M     & Compositional (3D shapes)                & Template & Open QA & \cmark & \xmark \\
CRIC \cite{gao2022cric}           & 96K    & 494K    & Commonsense (Daily life)       & Template & Open QA & \cmark & \xmark \\ 
AI2D \cite{hiippala2021ai2d}  & 5K    & 15K    & Scientific (Science diagram)                           & Manual & MCQA & \xmark & \xmark \\ 
ScienceQA \cite{saikh2022scienceqa}           & 10K    & 21K    & Scientific (Science problems)                           & Manual & MCQA & \xmark & \xmark \\ 
MMMU \cite{yue2024mmmu}     & 11.5K    & 11.5K    & Scientific (Exam questions)                           & Manual & MCQA & \xmark & \xmark \\ 
SEED-Bench \cite{li2024seed}           & 19K     & 19K     & Commonsense (Spatial, temporal)                & Neural & MCQA & \xmark & \xmark \\ 
MARVEL \cite{jiang2024marvel}      & 0.8K    & 3K    & Abstract shapes               & Template & Open QA & \xmark & \xmark \\ \midrule
\textbf{MultiStAR (Direct Answer)}               & 8.1K  & 21.7K  & \multicolumn{1}{c}{\begin{tabular}[c]{@{}c@{}}Abstract shapes\end{tabular}}
                                                                                & Template \& Neural & MCQA & \cmark & \xmark \\ 
\textbf{MultiStAR (Logical Chain)}               & 0.56K  & 3.92K  & \multicolumn{1}{c}{\begin{tabular}[c]{@{}c@{}}Abstract shapes\end{tabular}}
                                                                                & Template \& Neural & MCQA & \cmark & \cmark \\         
\bottomrule
\end{tabular}
}
\caption{Comparison of various VQA datasets. \textit{Template}: generated using predefined rules, \textit{Manual}: written by humans, \textit{Neural}: generated using large language models, \textit{Template \& Neural}: generated using predefined rules and rewritten by large language models. \textit{Open QA}: free-text answers, \textit{MCQA}: Multiple-Choice Question Answering. \textit{Functional Program}: Indicates whether the dataset is automatically created by functional programs. \textit{Multi-Step Structure}: Highlights whether the dataset includes a hierarchical structure with interdependent reasoning steps.}
\label{table:datasets}
\end{table*}

In summary, our contributions are: 1) we introduce the MultiStAR benchmark, designed to  evaluate models across different stages of reasoning through two subtasks, allowing for a more granular analysis of their performance throughout the reasoning process;
2) we present a novel metric that incorporates the correctness of the current stage as well as the accuracy of its dependent intermediate steps; and
3) we perform extensive experiments on a wide range of state-of-the-art MLLMs, our results indicate that current VLLMs still lack the ability to navigate multi-stage logical dependencies effectively. 

\begin{figure*}[ht]
    \centering
    \includegraphics[width=\textwidth]{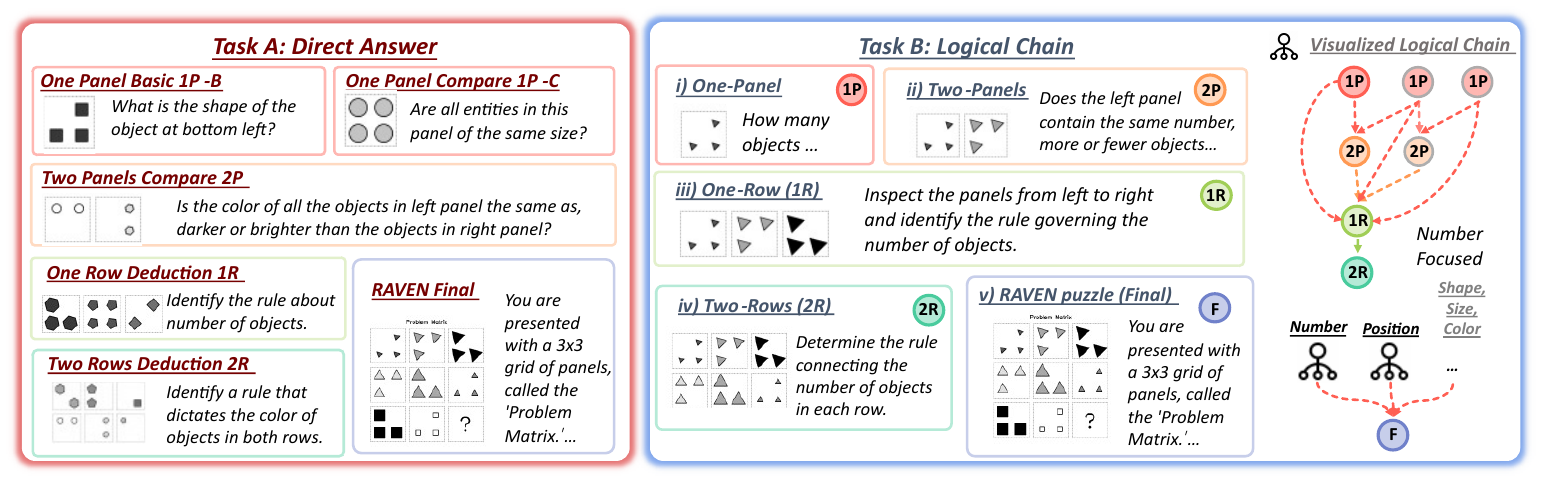}
    \caption{{Left:} Direct Answer subtask, showcasing six configurations along with their corresponding examples.
    {Right:} Logical Chain task, presenting a partial view of the logical chain (see full chain and the chain designing rationale in Appendix \ref{sec:The Full Logical Chain}). Examples are provided for \textbf{one specific path} in the chain.}
    \label{fig:task_define}
\end{figure*}

\section{Related Work}
Visual reasoning benchmarks have evolved to assess the capacity of AI models on various tasks including compositional \cite{johnson2017clevr}, commonsense \cite{gao2022cric,li2024seed}, scientific \cite{hiippala2021ai2d,saikh2022scienceqa,yue2024mmmu}, and abstract visual reasoning. 
Both commonsense and scientific reasoning tasks require real-world knowledge and a prior understanding of specific domains. Abstract Visual Reasoning (AVR) is the main focus of this work, and it primarily involves a classification task where models select an answer from a fixed set of choices based on abstract patterns and rules \cite{mandziuk2019deepiq,zhang2019raven,santoro2018measuring,nie2020bongard}. A few other AVR benchmarks address generative tasks, where models are tasked with recreating elements that fit within a given visual sequence, introducing additional complexity by evaluating a model’s creative reasoning capabilities \cite{chollet2019measure,moskvichev2023conceptarc}. The most similar benchmark to ours is MARVEL \cite{jiang2024marvel}, which targets AVR tasks and extends reasoning diversity with six core patterns across geometric and abstract shapes. It also includes basic perception questions to assess visual comprehension. However, MARVEL is still limited in its capacity to analyze intermediate reasoning steps.
Table \ref{table:datasets} shows the key statistics and features comparison of major multimodal reasoning datasets alongside our proposed MultiStAR benchmark.

\section{Multi-stage Evaluation Benchmark}
\subsection{Task Definition and Configuration}
Our dataset comprises two sub-tasks, Direct Answer and Logical Chain, both derived from RAVEN but with distinct reasoning patterns and focuses.
\subsubsection{Direct Answer}
To uncovering where the MLLMs likely to succeed or struggle \textit{in the individual stages}, this sub-task explores AVR across multiple levels, which is divided into six configurations, shown in Figure \ref{fig:task_define}:

 \noindent \textit{a) One Panel Basic Perception (\textbf{1P-B}):} The puzzle image consists of a single panel \( I = \mathbf{p} \), focusing on basic perception questions, such as determining the number of objects, the shape, or the position of a single object, without requiring any comparison.

     \noindent\textit{b) One Panel Comparison (\textbf{1P-C}):} The puzzle image remains a single panel, but questions require intra-panel attribute comparisons.

     \noindent \textit{c) Two Panels Comparison (\textbf{2P}):} The puzzle image consists of two panels, \( I = (\mathbf{p}_1 \), \( \mathbf{p}_2) \), requiring cross-panel comparisons. 

     \noindent \textit{d) One Row Rule Deduction (\textbf{1R}):} The puzzle image is a single row of three panels, \( I = (\mathbf{p}_1, \mathbf{p}_2, \mathbf{p}_3) \), and the task involves identifying a rule that governs the sequence. 

    \noindent  \textit{e) Two Row Rule Deduction (\textbf{2R}):} The puzzle image contains the first two rows, each with three panels, denoted as \( I = (\{\mathbf{p}_{1,1}, \mathbf{p}_{1,2}, \mathbf{p}_{1,3}\} \), \( \{\mathbf{p}_{2,1}, \mathbf{p}_{2,2}, \mathbf{p}_{2,3}\}) \). The task is to find a rule that applies to both rows.
    
    \noindent  \textit{f) RAVEN puzzle (\textbf{Final}):} The original puzzle from RAVEN dataset.
    
Formally, given an puzzle image \( I \) (which consist of one or more panels \( \mathbf{p} \)) and a question \( q \), the task is to select an answer \( a \) from a set of \( k \) multiple-choice options:

\begin{equation}
a^* = \arg\max_{a \in \mathcal{A}} P(a \mid I, q)
\end{equation}
where \( \mathcal{A} = \{a_1, \dots, a_k\} \) is the answer set. 
\subsubsection{Logical Chain}
To measure the \textit{sequential steps} of the reasoning process required to reach the final answer, rather than evaluating stages in isolation,
the second Logical Chain task extends reasoning across multiple subproblems, introducing dependencies between stages to form a coherent logical chain. As illustrated in the right part of Figure \ref{fig:task_define}, each node represents a stage question, and edges are connected if the previous information is necessary to answer the current stage. 

This task consists of five stages, similar to Direct Answer subtask: \textbf{\textit{1P}} (Merged 1P-B and 1P-C), \textbf{\textit{2P}}, \textbf{\textit{1R}}, \textbf{\textit{2R}} and \textbf{\textit{Final}}. 
Specifically, each node \( t \) involves predicting an answer \( a_t \) based on the current question \( q_t \), the current image \( I_t \), and information from prior stages \( \mathbf{H}_{t} \), the task is defined as:

\begin{equation}
a_t^* = \arg\max_{a_t \in \mathcal{A}_t} P(a_t \mid I_t, q_t, \mathbf{H}_{t})
\end{equation}
\begin{equation}
\mathbf{H}_{t} = \{\text{Re-Format}(q_j, a_j) \mid j \in \mathcal{D}_t\}
\label{eq:H_t}
\end{equation}
where \( \mathbf{H}_{t} \) represents the set of prior information, as determined by the pre-defined logical chain \( \mathcal{D}_t \), specifying one or more nodes that current node \( t \) depends on. As the images referenced by prior questions change across different stages, we use a rule-based program to reformat each dependent question 
\( q_j \) and the generated answer \( a_j \), appending this prior information before the current question to construct the input for the current node. Details of this program are provided in Appendix \ref{sec:Logical Chain}.


\begin{figure*}[ht]
    \centering
    \includegraphics[width=1.0\textwidth]{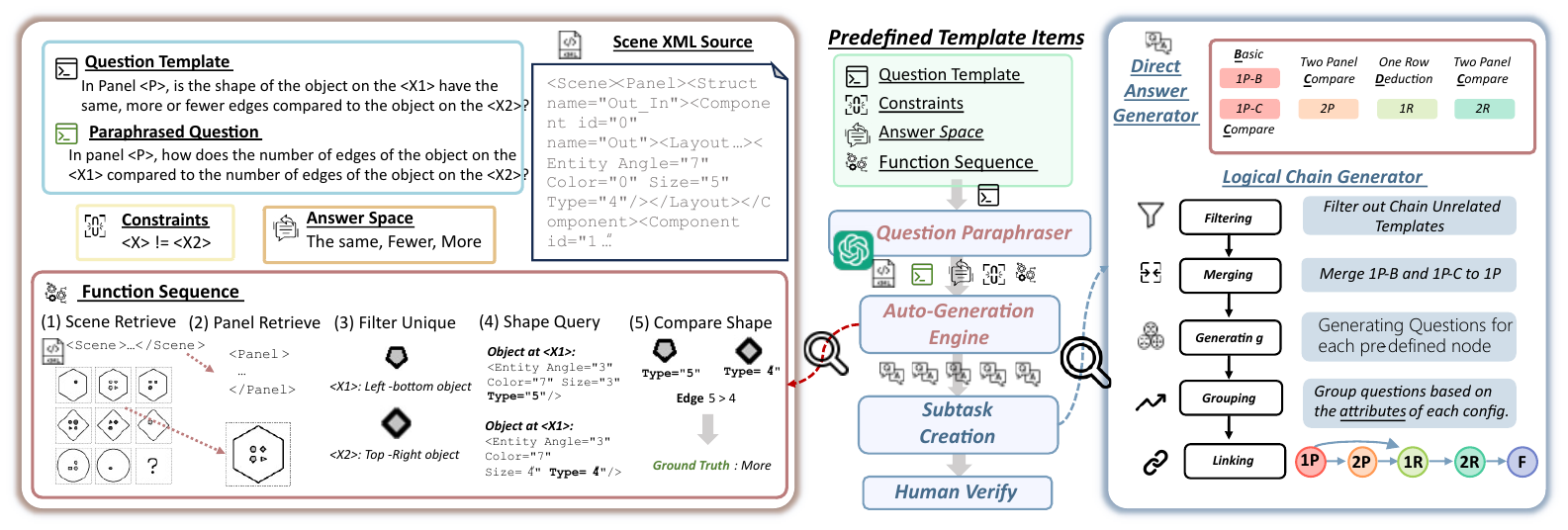}
    \caption{Our MultiStAR dataset generation pipeline.}
    \label{fig:dataset_creation}
\end{figure*}

\subsection{Dataset Creation}
\label{sec:Dataset Creation}
\noindent\textbf{Data Sources:} Our MultiStAR dataset is derived from the RAVEN dataset, which its associated XML files provide objects details and ground-truth logical rules for generating each puzzle. 

\noindent \textbf{Define Templates}: We pre-define question templates for all six configurations, each template including a question format, constraints, an answer space, and a corresponding function sequence, as illustrated in Figure \ref{fig:dataset_creation}. Overall, we created 25 distinct templates, details are shown in Appendix \ref{sec:Generation Template}. 

\noindent\textbf{Question Generation}: By leveraging the puzzle information and the pre-defined templates, we implement an automated template-based generation process to efficiently produce large-scale question-answer pairs. Firstly, to enrich question formats and linguistic diversity, we employ GPT-4o \cite{gpt4o} to rewrite the templates. Then, following a methodology similar to the CLEVR dataset \cite{johnson2017clevr}, we design functional programs that execute a sequence of functions. For instance, as shown in Figure \ref{fig:dataset_creation}, the program ``Scene Retrieve → Panel Retrieve → Filter Unique → Shape Query → Compare Shape'' identifies the puzzle matrix, retrieves the relevant panel <P>, locates objects at positions <X1> and <X2>, queries their shapes, and compares them to determine the ground-truth answer. And lastly, the multiple-choice options are sampled from the answer space. 

\noindent\textbf{Subtasks Creation:} To create the Direct Answer subtask, we first sample XML files from RAVEN, and for each XML file, we generate one question for each \textit{template}. During question formation, placeholders (e.g., <X1>, <X2>) are replaced with randomly selected any possible values consistent with the value ranges and constraints. Next, we create the Logical Chain sub-task by first filtering out those templates that do not contribute useful information for building the logical chain (i.e. the question does not provide necessary input for its child nodes). To simplify chain construction, the first two one-panel configurations, 1P-B and 1P-C, are combined into a single stage representing one-panel information. During question generation, one question is created for each \textit{node}, with placeholders such as panel 
<P> replaced by the values corresponding to the current node's position in the chain. For instance, if there are three 1P nodes in the chain, they correspond to panels 1, 2, and 3, respectively. Questions are then grouped by attributes such as number and position, aligning with how the chain is constructed. Finally, we assign the previous nodes for each question to establish the edges between nodes. 
Detailed analysis of our dataset MultiStAR, such as the question length and function distribution, please see Appendix \ref{sec:Dataset Analysis}.

\noindent\textbf{Human Verification}: To evaluate the quality of automatically generated question-answer pairs, we also conduct human study based on three aspects, \textit{Correctness}, \textit{Clarity} and \textit{Content Validity}. The results show our dataset performing well across all aspects, see Appendix \ref{sec:Part A} for details.

\subsection{Evaluation Metrics}
We use accuracy for the Direct Answer subtask, as it directly aligns with the task of selecting the correct answer from multiple choices. 
However, for the \textbf{Logical Chain} subtask, accuracy alone does not consider intermediate reasoning steps, focusing only on the end result. To address this limitation and better align with the step-by-step reasoning process, we introduce a new metric, MSEval. As the example illustrated in Figure \ref{fig:metrics}, the score for the 1R node is designed to aggregate from all its related nodes, which include three 1P nodes, two 2P nodes, and the 1R node itself. This aggregation captures the interconnectivity between nodes in the logical chain. And to reflect their contribution to the reasoning process, it assigns a weight to each of these nodes based on their importance. Specifically:
\begin{figure}[ht]
    \centering
    \includegraphics[width=0.49\textwidth]{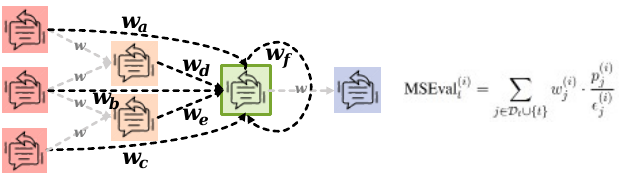}
    \caption{This example is the MSEval score calculation for the \textcolor{lowcontrastgreen}{1R} node, depends on \textcolor{lowcontrastred}{1P}, \textcolor{lowcontrastyellow}{2P} and \textcolor{lowcontrastgreen}{1R} itself. The corresponding weights are denoted as \(w_a\) through \(w_f\).}
    \label{fig:metrics}
\end{figure}

\noindent\textbf{Aggregated Outcomes:} 
To aggregated the intermediate steps outcomes across the chain, MSEval chooses to compute a joint probability of the current node and all dependent nodes as the product of their conditional probabilities. To prevent disregarding the model's performance when it is incorrect, each conditional probability is derived by measuring the probability assigned to the ground truth. Using logits from the model’s final layer for the answer choices (e.g., A, B, C, D), the correct answer logit \( p_j^{(i)} \) is transformed into a probability via the softmax function. This is defined as:

\begin{align}
\text{Joint Prob}_t^{(i)} &= \text{Norm}(P(a_t^{(i)} = a_t^{(i)*}, \Phi_t^{(i)} \mid \xi_t^{(i)})) \notag \\
&= \prod_{j \in \mathcal{D}_t \cup \{t\}} \exp(\frac{p_j^{(i)}}{\epsilon_j^{(i)}})
\end{align}
\begin{equation}
p_j^{(i)} = P(a_j^{(i)} = a_j^{(i)*} \mid \mathbf{H}_j^{(i)}) = \frac{\exp(z_{a_j^{(i)*}})}{\sum_{k \in \mathcal{A}_j^{(i)}} \exp(z_k)}
\end{equation}
%
where \( z_{a_j^{(i)*}} \) is the logit for the correct answer \( a_j^{(i)*} \), and \( \mathcal{A}_j^{(i)} \) is the set of all possible answers, \( \epsilon_j^{(i)} = \frac{1}{|\mathcal{A}_j^{(i)}|} \) represents the random rate. The term \( \exp(\cdot) \) normalizes the probability to account for varying numbers of choices for all nodes. \(\Phi_t^{(i)} = \{a_j^{(i)} = a_j^{(i)*} \mid j \in \mathcal{D}_t\}\) represents the correct answer probability for all dependent nodes, \( \xi_t^{(i)} = \{\mathbf{H}_j^{(i)} \mid j \in \mathcal{D}_t \cup \{t\}\}\) represents a set of prior information for each node.

\noindent\textbf{Weighted Importance}: The joint probability does not account for the relative importance of each node in the chain. 
To address this, we introduce a weight for each node based on its influence on the current node, as measured by conditional mutual information (CMI). 
CMI is obtained by altering the set of all possible answers \( \mathcal{A}_j^{(i)} \) at node \( j \) while keeping the outputs of all other nodes (\( \mathcal{D}_t^{(i)} \setminus \{j\} \)) fixed. We then observe how the model's outputs \( \mathcal{A}_{j \to t}^{(i)} \) at the current node \( t \) change. If \( \mathcal{A}_{j \to t}^{(i)} \) changes significantly, the CMI is higher, resulting in a higher weight. As raw CMI values may vary in scale, we take a normalization of the conditional mutual information (NCMI).
This process is defined as:

\begin{align}
\text{CMI}(i, j, t) &= \text{CMI}(\mathcal{A}_j^{(i)}; \mathcal{A}_{j \to t}^{(i)} \mid \mathcal{D}_t^{(i)} \setminus \{j\}) \notag \\ &= H(\mathcal{A}_{j \to t}^{(i)} \mid \mathcal{D}_t^{(i)} \setminus \{j\}) \nonumber \\
&\quad + H(\mathcal{A}_j^{(i)} \mid \mathcal{D}_t^{(i)} \setminus \{j\}) \nonumber \\
&\quad - H(\mathcal{A}_{j \to t}^{(i)}, A_j^{(i)} \mid \mathcal{D}_t^{(i)} \setminus \{j\})
\end{align}
\begin{equation}
\text{NCMI}(i, j, t) = \frac{\exp(\text{CMI}(i, j, t))}{\sum_{k \in \mathcal{D}_t \cup \{t\}} \exp(\text{CMI}(i, k, t)))}
\end{equation}
%
where \( H (\cdot) \) denotes entropy. Note that for current node \(t\), we have \( \mathcal{A}_t^{(i)} = \mathcal{A}_{t \to t}^{(i)} \), so that current node would always have the highest impact to itself.


We apply the NCMI to each node's conditional probability to compute its weighted contribution to the reasoning process. To simplify the formulation, we apply a log to the expression. The final \textbf{MSEval} score for stage \( t \), instance \( i \) is computed as:

\begin{align}
\text{MSEval}_t^{(i)} &= \log \prod_{j \in \mathcal{D}_t \cup \{t\}} (\exp(\frac{p_j^{(i)}}{\epsilon_j^{(i)}}))^{\text{NCMI}(i, j, t)} \notag \\
&= \sum_{j \in \mathcal{D}_t \cup \{t\}} \text{NCMI}(i, j, t) \cdot \frac{p_j^{(i)}}{\epsilon_j^{(i)}}
\end{align}
\begin{equation}
\text{MSEval}_t^{(i)} = \sum_{j \in \mathcal{D}_t \cup \{t\}} w_j^{(i)} \cdot \frac{p_j^{(i)}}{\epsilon_j^{(i)}}
\end{equation}

As MSEval relies on access to the logits of the model's final layer, which can only be applied to \textbf{open-source} models. More details about MSEval, such as algorithm Pseudo Code and computational cost, are shown in Appendix \ref{sec:Additional Details about MSEval}.

\section{Results}
In the Direct Answer subtask, we evaluate 17 representative MLLMs in zero-shot setting, including both open-source and close-source models. For open-source models, we have vanilla pre-trained (i.e., without instruction-tuning) and instruction-tuned models. Additionally, we consider a range of model sizes to ensure a comprehensive evaluation. The model settings and the input prompt details are shown in Appendix \ref{sec:Direct Answer}.

In the Logical Chain subtask, we evaluate six models which performed well in the Direct Answer subtask in zero-shot setting. To compare how much benefit the models gain from prior information, we also evaluate the models without providing prior information (\( \mathcal{\mathbf{H}}_t = \emptyset \)), serving as a baseline for comparison. In this case, as we only consider the current stage, the MSEval score simplifies to \(\text{MSEval}_t^{(i)} = \frac{p_t^{(i)}}{|\mathcal{A}_t^{(i)}|} \). The model settings and the input prompt details are shown in Appendix \ref{sec:Logical Chain}.

To establish the human performance on the sub-problems within MultiStAR, we conduct a human study on a crowd-sourcing platform in which participants solved a 10\% subset of the benchmark. We do not evaluate human performance on the original RAVEN puzzles, but instead use the result reported in \citet{zhang2019raven}. Please see Appendix \ref{sec:Part B} for more details about the human study.

\subsection{Result Analysis}
\paragraph{Direct Answer:}Table \ref{table:answer_result} compares the performance of various MLLMs on the Direct Answer task. The two close-source models, GPT-4o and Gemini-1.5-pro, outperform others in 1P-B, 1R and 2R stages. GPT-4o achieves an impressive 88.07\% accuracy for basic object-oriented questions within a single panel, highlighting its strong capability to recognize simple visual patterns. Gemini achieves the best results in rule deduction tasks for both one-row and two-row configurations, indicating its superior ability to process more complex visual inputs and perform logical reasoning effectively. Among open-source models, pre-trained models generally perform worse than instruction-tuned ones, which is no surprise as the pre-trained models are unlikely to be able follow the prompt/instructions. Qwen2-VL-Instruct-72B achieves the best performance on cross-panel comparison tasks. Interestingly, we also see that it has a very strong performance in the original RAVEN task (``Final''), and we suspect there is probably contamination since it struggles with the more complex intermediate reasoning tasks (``1R'' and ``2R''). We also think Idefics2-8B maybe be contaminated, again for the same reason.

We performed an error analysis (see Appendix \ref{sec:error analysis}) and identified that the models mainly face challenges with deep reasoning errors. Additionally, the decoder-only models (Qwen2-VL, InternVL2 and NVLM-D) tend to perform much better than encoder-decoder models (InstructBLIP and LLaVA-v1.5). This suggests that decoder-only architectures may be more effective for step-by-step reasoning to some extent. However, it's important to note that the decoder-only models are newer and the performance gap may be attributed to other factors such as larger or more diverse training data or better training objectives.

Interestingly, as the questions become increasingly complex and require deeper reasoning, a noticeable decline in performance is observed across all models, gradually approaching the random baseline. While models demonstrate strong performance on basic perception tasks, they struggle significantly with deeper reasoning challenges. In contrast, human performance remains stable at above 60\% with increasing complexity. This highlights the substantial gap between model and human performance, emphasizing the limitations of current MLLMs in understanding and reasoning at a level comparable to humans. 
Another finding is that performance generally improves with larger model sizes --- the size of the base language model (see Appendix \ref{sec:Model Parameters Trend} for detailed visual analysis). This suggests that these VLLMs might be primarily relying on its language model for tackling these AVR tasks.
\begin{table}[ht]
\setlength{\tabcolsep}{4pt}
\renewcommand{\arraystretch}{1.1}
\small
\resizebox{0.485\textwidth}{!}{
\begin{tabular}{lcccccc}
\hlineB{4}
                                         & 1P-B 
                                             & 1P-C 
                                             & 2P 
                                             & 1R 
                                             & 2R 
                                             & Final        \\ \hline\hline
\multicolumn{7}{l}{\textbf{Close-Source Models}} \\ \hline                                    
                                    
GPT-4o \shortcite{gpt4o}              & \textbf{88.1}           & 72.7                      & 54.0                      & 40.0            & 31.6            & 12.1 \\
Gemini-1.5-pro \shortcite{team2023gemini}                          & 83.2                    & 75.0                      & 50.0                      & \textbf{46.9}   & \textbf{37.8}   & 11.6 \\ \hline\hline
\multicolumn{7}{l}{\textbf{Pre-trained Open-Source Models}} \\ \hline 
 
Qwen-VL-7B \shortcite{bai2023qwen}                     & 17.5                    & 24.7                      & 22.5                      & 15.8            & 12.2            & 12.3 \\
Idefics2-8B \shortcite{laurenccon2024matters}            & 17.2                    & 33.0                      & 27.3                      & 19.8            & 21.4            & 12.3 \\
xGen-MM-4B \shortcite{xue2024xgen}                & 40.2                    & 31.8                      & 12.5                      & 24.1            & 23.9            & 3.4  \\ \hline \hline
\multicolumn{7}{l}{\textbf{Instruction-Tuned Open-Source Models}} \\ \hline

Instructblip-7B \shortcite{dai2023instructblipgeneralpurposevisionlanguagemodels}   & 27.5                    & 37.1                      & 27.7                      & 14.0            & 13.5            & 11.6 \\
Instructblip-13B \shortcite{dai2023instructblipgeneralpurposevisionlanguagemodels} & 29.4                    & 39.0                      & 26.9                      & 25.0            & 23.0            & 14.3 \\
LLaVA-v1.5-7B \shortcite{liu2024visual}           & 47.8                    & 50.2                      & 32.9                      & 27.1            & 25.7            & 13.3 \\
LLaVA-v1.5-13B \shortcite{liu2024visual}          & 59.6                    & 47.6                      & 15.9                      & 26.7            & 26.9            & 11.3 \\
Idefics2-8B$^\ast$ \shortcite{laurenccon2024matters}             & 85.1                    & 65.5                      & 42.0                      & 37.0            & 36.8            & 29.9 \\
xGen-MM-4B \shortcite{xue2024xgen}           & 81.2                    & 47.9                      & 21.9                      & 24.0            & 25.6            & 2.4  \\
Qwen2-VL-2B \shortcite{wang2024qwen2}     & 40.4                    & 42.7                      & 22.8                      & 13.7            & 11.4            & 9.9\\
Qwen2-VL-7B$^\ast$ \shortcite{wang2024qwen2}    & 64.8                    & 56.6                      & 47.9                      & 31.0            & 33.2            & 24.3\\
Qwen2-VL-72B$^\ast$ \shortcite{wang2024qwen2}   & 86.9                    & \textbf{77.8}             & \textbf{60.2}             & 45.5            & 21.9            & \textbf{63.7}  \\
NVLM-D-72B \shortcite{dai2024nvlm}              & 80.5                    & 67.1                      & 45.3                      & 39.1            & 31.3            & 12.7 \\
Intern-VL2-2B \shortcite{chen2024far}                               & 54.0                    & 48.7                      & 27.3                      & 26.9            & 23.9            & 10.1 \\
Intern-VL2-8B \shortcite{chen2024far}                               & 63.0                    & 54.5                      & 34.2                      & 23.2            & 23.4            & 14.6 \\ 
\hline\hline
Random                                      & 39.9                    & 23.0                      & 26.2                      & 25.0            & 25.0            & 12.5    \\
Human                                      & 98.5                    & 88.9                      & 69.1                      & 62.1            & 63.3            & 84.4$^\ddagger$    \\
\hlineB{4}
\end{tabular}
}
\caption{The answer accuracy of MLLMs for the Direct Answer subtask. The best results are highlighted in bold. $^\ast$ denotes contaminated models that may have been trained with RAVEN. $^\ddagger$ means the number is taken from from  \citet{zhang2019raven}.}
\label{table:answer_result}
\end{table}

\begin{table}[ht]
\setlength{\tabcolsep}{4pt}
\renewcommand{\arraystretch}{1.15}
\small
\resizebox{0.48\textwidth}{!}{
\begin{tabular}{lccccccc}
\hlineB{4}  
                                & Metric                      & Prior & 1P     & 2P     & 1R     & 2R     & Final       \\ \hline\hline
\multirow{2}{*}{GPT-4o}         & \multirow{2}{*}{Acc}   & w/o        & 73.8   & 39.1   & 34.7   & 28.9   & 15.7        \\
                                &                            & w          & 73.8   & 43.9   & 41.8   & 50.6   & 10.0        \\ \cline{1-8}                               
\multirow{2}{*}{Gemini-1.5-pro}         & \multirow{2}{*}{Acc}   & w/o        & 75.5   & 61.6   & 49.6   & 44.6   & 5.7         \\
                                &                            & w          & \textbf{75.5} & \textbf{64.4} & \textbf{52.6} & \textbf{57.1} & 18.6 \\ \cline{1-8}                               
\multirow{4}{*}{\makecell{Idefics2-8B$^\ast$}}    & \multirow{2}{*}{Acc}   & w/o        & 57.8   & 39.6   & 34.4   & 35.1   & 25.7        \\
                                &                            & w          & 57.8   & 37.8   & 36.6   & 42.4   & 25.7        \\ \cline{2-8}
                                & \multirow{2}{*}{MSEval}     & w/o        & 2.02   & 1.29   & 1.29   & 1.27   & 1.24        \\
                                &                            & w          & 2.02   & 1.48   & 1.51   & 1.51   & 1.44        \\ \cline{1-8}
\multirow{4}{*}{\makecell{Qwen2-VL-72B$^\ast$}}   & \multirow{2}{*}{Acc}   & w/o        & 74.1   & 56.0   & 45.1   & 42.9   & 64.3       \\
                                &                            & w          & 74.1   & 57.8   & 47.3   & 54.2   & \textbf{65.7} \\ \cline{2-8}
                                & \multirow{2}{*}{MSEval}     & w/o        & 2.54   & 1.95   & 1.79   & 1.70   & \underline{5.14} \\
                                &                            & w          & \underline{2.54} & \underline{2.13} & \underline{2.12} & \underline{2.10} & 3.31 \\ \cline{1-8}
\multirow{4}{*}{\makecell{Intern-VL2-8B}} & \multirow{2}{*}{Acc}   & w/o        & 54.4   & 35.9   & 21.6   & 21.6   & 18.6        \\
                                &                            & w          & 54.4   & 41.9   & 31.6   & 33.5   & 17.1        \\ \cline{2-8}
                                & \multirow{2}{*}{MSEval}     & w/o        & 1.75   & 1.09   & 0.90   & 0.91   & 1.18        \\
                                &                            & w          & 1.75   & 1.38   & 1.30   & 1.18   & 1.26        \\ \cline{1-8}
\multirow{4}{*}{\makecell{NVLM-D-72B}}     & \multirow{2}{*}{Acc}   & w/o        & 66.1   & 42.4   & 37.8   & 23.5   & 8.6         \\
                                &                            & w          & 66.1   & 45.2   & 39.1   & 43.3   & 7.1         \\ \cline{2-8}
                                & \multirow{2}{*}{MSEval}     & w/o        & 2.25   & 1.20   & 1.28   & 1.02   & 0.76        \\
                                &                            & w          & 2.25   & 1.65   & 1.69   & 1.62   & 1.41        \\ \cline{1-8}
\multirow{2}{*}{Random}         & Acc   & -          & 31.1   & 31.7   & 25.0   & 25.0   & 12.5        \\
                                & MSEval     & -          & 1.00   & 1.00   & 1.00   & 1.00   & 1.00        \\ \hlineB{4}
\end{tabular}
}
\caption{The Accuracy (Acc) and MSEval scores for the Logical Chain task. $^\ast$ denotes contaminated models that may have been trained with RAVEN. The highest accuracy are highlighted in \textbf{bold}. The highest MSEval are highlighted in \underline{underline}. w/o: without prior, w: with prior. See Appendix \ref{sec:Logical Chain} for results of more models.}
\label{table:merged}
\end{table}

\paragraph{Logical Chain:}
Table \ref{table:merged} shows the performance of MLLMs on each stage of the Logical Chain task. Results are reported as accuracy and MSEval scores. Among all models, Gemini-1.5-pro achieves the best performance on the first four stages. Among the open-source models, Qwen2-VL-72B outperforms others in both accuracy and MSEval, although as mentioned earlier we think the model has been trained with RAVEN and we should probably interpret these results with caution (same for Idefics2-8B).
When comparing results with and without prior information, all models show better performance when prior information is available in both metrics, highlighting their capacity to benefit from step-by-step reasoning, even when some generated previous answers might be incorrect. For visual representation of the percentage increase, see to Appendix \ref{sec:Performance Increase with Prior Info}.

Interestingly, for the final stage involving the RAVEN puzzle, prior information appears to provide limited utility for most models (exception: Gemini). This aligns with previous findings that chain-of-thought reasoning models struggle to solve RAVEN puzzles \cite{ahrabian2024curious, gendron2024large}. However, MSEval scores tell a different story. Almost all models (exception: Qwen2-VL-72B) see an improvement in MSEval even though there is no change (or even a drop) in accuracy when prior information is incorporated. 
Since MSEval evaluates correctness across intermediate and current stages, it reveals that even though prior information might not help MLLMs solve the final stage question, it helps them with the intermediate tasks.

Another finding is the MSEval score for Qwen2-VL-72B declines when provided with prior information, which is not consistent with accuracy. This indicates its weakness in addressing intermediate stages that were likely not part of its pre-training. Despite errors in earlier stages, the model still performs well on the final question, suggesting it relies on memorizing patterns from the final stage rather than demonstrating a strong understanding of the logical reasoning behind the task. This highlights a critical limitation in current MLLMs, \textbf{while they may achieve impressive results in isolated cases, their ability to generalize and reason through multi-stage logical dependencies remains inadequate}. 
To further verify the MSEval's effectiveness, we also conduct qualitative analysis in section \ref{sec:Qualitative Analysis Main}.
\section{Discussion}

\begin{figure}[ht]
    \centering
    \includegraphics[width=0.50\textwidth]{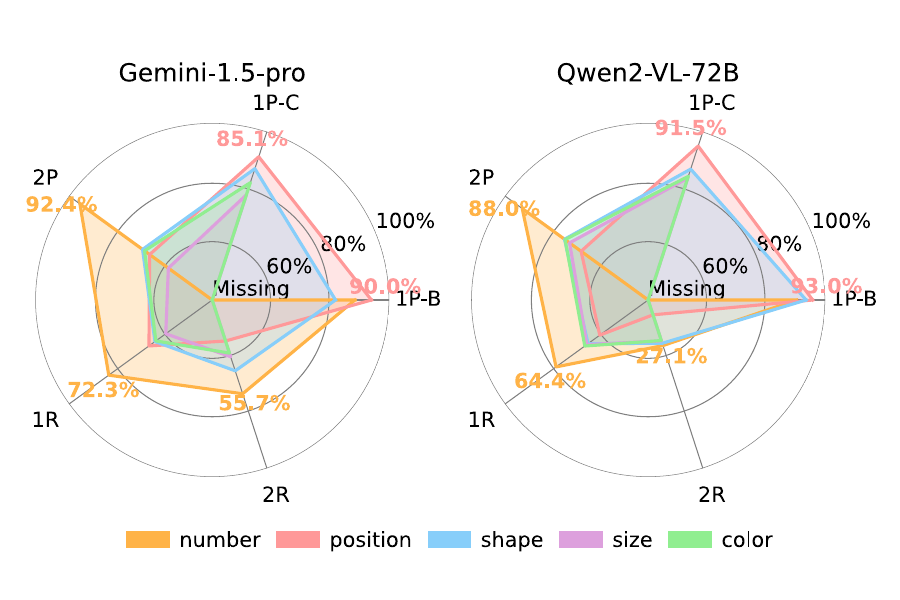}
    \caption{Breakdown analysis of five attributes for Gemeni-1.5-pro and Qwen2-VL-72B on the Direct Answer task. See Appendix \ref{sec:Attribute Break-Down Analysis} for more models.}
    \label{fig:radar}
\end{figure}

\paragraph{What insights can be drawn from each attribute's performance?} From Figure \ref{fig:radar}, the ``number'' attribute is the easiest to recognize in higher level configurations (2P, 1R, 2R), while ``position'' is the most easily identified in low-level, single-panel settings. Some attributes achieve accuracy above 90\%, indicating that the models exhibit strong counting and spatial reasoning capabilities. However, they struggle with attributes like ``color'' and ``size'', particularly in high-level configurations.

\paragraph{Given the ground truth (i.e., correct) answer  for intermediate steps, how does it influence the final results?} Table \ref{table:ground truth} highlights that the correctness of the prior information is important. For example, the 1R stage benefits significantly from the insights about each panel and intra-panel comparisons. The 2R stage also sees substantial gains, as it mainly relies on double-checking information from the 1R stage without requiring additional changes in most cases. However, the final stage experiences a negative impact despite the inclusion of correct rules. This may be attributed to the complexity of the visual input, which contains numerous objects, making it challenging for the model to effectively apply the given rules. And for the Qwen2-VL-72B model, its tendency to memorize patterns might turn these ground truths into noise. 

\begin{table}[ht]
\setlength{\tabcolsep}{4.5pt} 
\renewcommand{\arraystretch}{1.2} 
\small
\centering
\resizebox{0.48\textwidth}{!}{ 
\begin{tabular}{lcccccc}
\hlineB{4}
&            Prior Info  & 1P & 2P & 1R & 2R & Final \\ \hline \hline
\multirow{2}{*}{GPT-4o}          & w/o   & 73.8     & 39.1      & 34.7   & 28.9    & 15.7   \\
 & GT & 73.8 &	49.2 &	66.0 &	93.8 &	14.3    \\ \cline{2-7}
\multirow{2}{*}{Gemini-1.5-pro}       & w/o   & 75.5     & 61.6      & 49.6   & 44.6    & 5.7   \\
& GT & \textbf{75.5}	& \textbf{62.4} &	68.9 &	79.5 &	7.1   \\\cline{2-7}
\multirow{2}{*}{\makecell{Qwen2-VL-72B$^\ast$}}   & w/o   & 74.1     & 56.0      & 45.1   & 42.9    & 64.3   \\
& GT & 74.1 &	61.9 &	\textbf{70.7} &	92.2 &	\textbf{55.7}   \\\cline{2-7}
\multirow{2}{*}{\makecell{NVLM-D-72B}}     & w/o   & 66.1     & 42.4      & 37.8   & 23.5    & 8.6   \\
& GT  & 66.1 &	51.9 &	66.4 &	\textbf{96.0} &	7.14\\ \hlineB{4}
\end{tabular}
}
\caption{The accuracy with incorporating ground truth information at each stage for Logical Chain task.}
\label{table:ground truth}
\end{table}

\paragraph{How much do previous stages influence the current stage?} 
A key step in our MSEval metrics is measuring the relative importance of intermediate dependent stages to the current stage using NCMI. This allows us to assess how each step in the chain depends on prior stages, \textit{helping verify whether previous information is useful and if the designed chain is logically sound}. As shown in Table \ref{table:influence}, prior information often has significant weight on the current stage, except for the position attribute at ``1P to 2P''. This suggests that querying object position in a single panel has little impact on determining if positions are the same across panels.

\begin{table}[ht]
\setlength{\tabcolsep}{4pt}
\renewcommand{\arraystretch}{1.2}
\small
\centering
\resizebox{0.48\textwidth}{!}{
\begin{tabular}{lccccc}
\hlineB{4}
                                & Attributes          & 1P to 2P    & (1P,2P) to 1R     & 1R to 2R    & 2R to F    \\ \hline \hline
\multirow{5}{*}{\makecell{Qwen2-VL-72B}}         & Number         & 0.40   & 0.54   & 0.33   & \multirow{5}{*}{\centering 0.42}         \\
                                & Position       & 0.21	& 0.48   &	0.39   &            \\
                                & Shape        & 0.37 &	0.51   &	0.27   &             \\ 
                                & Color        & 0.36 &	0.53   &	0.33   &             \\ 
                                & Size        & 0.37 &	0.51   & 0.32   &             \\ \cline{1-6}                               
\multirow{5}{*}{\makecell{NVLM-D-72B}}         & Number         & 0.40   & 0.56  & 0.37   & \multirow{5}{*}{\centering 0.63}         \\
                                & Position       & 0.21	& 0.50   &	0.42   &            \\
                                & Shape        & 0.40 &	0.52   &	0.35   &             \\ 
                                & Color        & 0.38 &	0.54   & 0.38   &             \\ 
                                & Size        & 0.39 &	0.54   &	0.38   &             \\ \hlineB{4}
\end{tabular}
}
\caption{Average NCMI weight assigned to all dependent stages, grouped by each attribute.}
\label{table:influence}
\end{table}

\begin{figure*}[ht]
    \centering
    \includegraphics[width=0.8\textwidth]{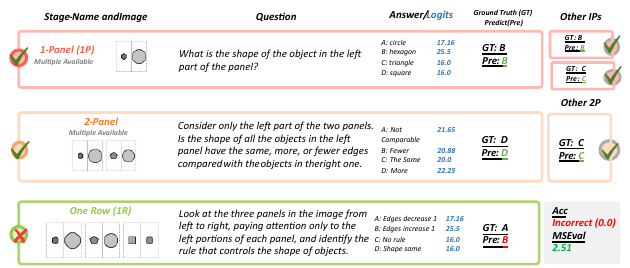}
    \caption{The top two rows are dependent stages (All Correct), the bottom row is current stage (Incorrect).}
    \label{fig:qualititave_analysis}
\end{figure*}

\paragraph{How do variations in handling long prompts affect model outcomes?} Injecting prior information into prompts significantly increases their length (see Appendix \ref{sec:Dataset Analysis} for details), making it more challenging for models to focus on critical details. To address this issue, we proposed two methods: (1) adding HTML tags to structure the prompt by separating prior information, background, and questions, enabling the model to clearly distinguish each part, and (2) formatting the prompt as a PDF document with distinct sections and titles. Table \ref{table:long prompt} demonstrates that HTML tagging provides some benefits, but not so much for the PDF approach (see Appendix \ref{sec:Handling Long Prompts} for further details and examples of the conversion methods).

\begin{table}[ht]
\setlength{\tabcolsep}{4pt}
\renewcommand{\arraystretch}{1.2}
\small
\centering
\resizebox{0.48\textwidth}{!}{
\begin{tabular}{lcccccc}
\hlineB{4}
                                & Prior          & 1P     & 2P     & 1R     & 2R     & Final       \\ \hline\hline
\multirow{3}{*}{GPT-4o}         & Vanilla         & 73.8   & 43.9   & 41.8   & 50.6   & 10.0        \\
                                & Struct.       & \textbf{82.2}	& 64.4 &	47.8 &	50.9 &	8.6           \\
                                & Doc.        & 80.8 &	44.8 &	31.1 &	24.9 &	10.0           \\ \cline{1-7}                               
\multirow{3}{*}{Gemini-1.5-pro}         & Vanilla         & 75.5 & 64.4 & 52.6 & 57.1 & 18.6 \\
                                & Struct.       & 70.6	& 66.4 &	52.9 &	\textbf{57.8} &	17.1           \\
                                & Doc.        & 69.6 & 	51.0 &	36.7 &	33.1 &	14.3              \\ \hline
\multirow{3}{*}{Qwen2-VL-72B$^\ast$}         & Vanilla         & 74.1	& 57.8	& 47.3	& 54.2	& 65.7 \\
                                & Struct.       & 77.2	& \textbf{67.7}& 	\textbf{55.1}	& 53.6	& 61.4           \\
                                & Doc.        & 76.5	& 63.1& 	50.2	& 46.6& 	24.3              \\ \hlineB{4}
\end{tabular}
}
\caption{The accuracy of three prompting techniques for prior information for Logical Chain task. \textit{Vanilla}: Pure Text, \textit{Struct.}: Structure (HTML), \textit{Doc.}: Document.}
\label{table:long prompt}
\end{table}

\section{Qualitative Analysis}
\label{sec:Qualitative Analysis Main}

To highlight the advantages of our MSEval metric over traditional accuracy, we provide several concrete examples across different scenarios. Figure \ref{fig:qualititave_analysis} shows a case where the final answer to the current question is incorrect, resulting in an accuracy score of 0.0. However, the model demonstrates strong performance in intermediate steps, correctly solving the one-panel and two-panel comparisons with high confidence. Additionally, the logits for the correct answer "A" are only slightly lower than the highest logits. By considering these factors, the MSEval metric produces a more reasonable score, reflecting the model's partial success. Further examples, including cases where the current question is correct but the intermediate steps are not, are provided in Appendix \ref{sec:Qualitative Analysis}. 


\section{Conclusion}
In this work, we propose the MultiStAR benchmark,  to evaluate MLLMs on the intermediate steps in the reasoning process through two tasks, Direct Answer and Logical Chain. While current models perform well on basic perception tasks, they face significant challenges with deeper reasoning stages. Our findings also indicate that models have the potential to benefit from step-by-step reasoning. However, despite extensive training yielding impressive results in isolated scenarios, their ability to handle logical dependencies remains limited. We introduce a metric MSEval, which can be applied to a variety of reasoning tasks beyond visual reasoning, including domains such as mathematics and science, where multi-step logic is critical, provided there are clearly defined chains. 
\section{Limitation}
The intermediate task generation method we propose is restricted to datasets with explicitly defined object attributes (in our case, they are documented in the XML files in RAVEN). This limits its applicability to other AVR datasets, as they often lack such metadata.

The logical chain design in our dataset is not perfect. In some cases, prior information is insufficient for the current stage, such as instances in the one-row rule deduction stage where the rule might involve ``Three Different Numbers'', in this case, we also need the information about the second row. To make the chain construction easier, we design chains at the ``Corpus-Level'', meaning they are fixed across all instances. Future work could explore automatic ``Instance-Level'' chain construction methods to enable models to dynamically generate chains based on patterns within individual examples.


\section*{Acknowledgments}

This project is supported by the Australian Research Council Discovery Grant awarded to Jey Han Lau (ID: DP240101006).


\clearpage

\begin{figure*}[ht]
    \centering
    \hspace{-1.0em}
    \begin{subfigure}{0.24\textwidth}
        \centering
        \includegraphics[height=3.4cm]{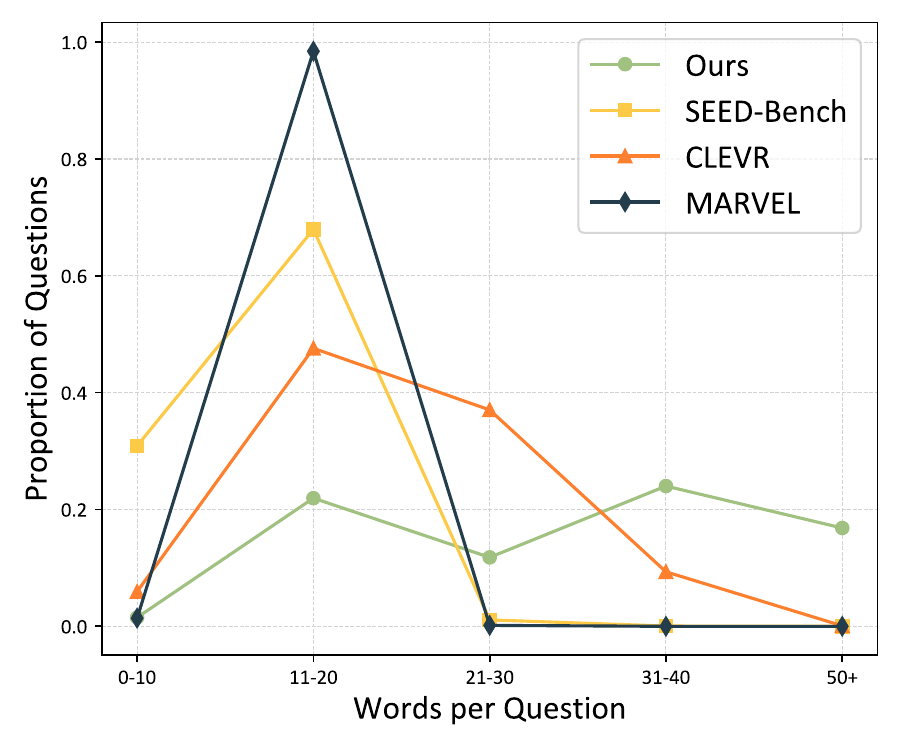}
        \caption{Question Length Compare}
        \label{fig:analysis_1}
    \end{subfigure}
    \hspace{+0.7em}
    \begin{subfigure}{0.24\textwidth}
        \centering
        \includegraphics[height=3.54cm]{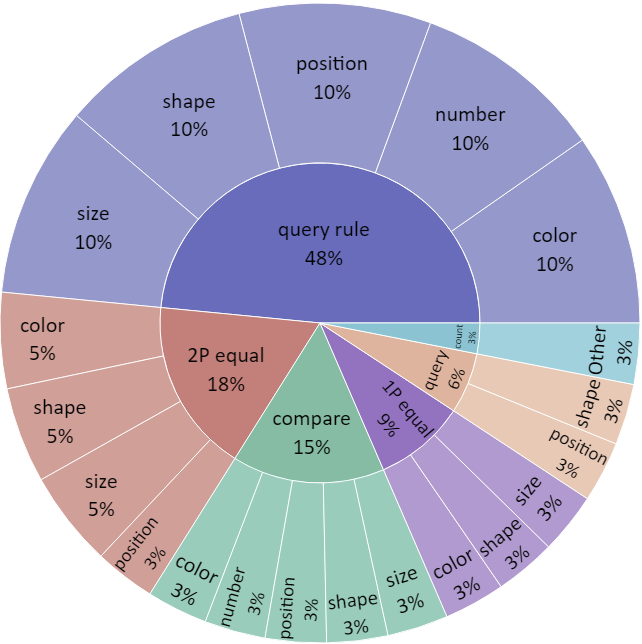}
        \caption{Function Distribution}
        \label{fig:analysis_2}
    \end{subfigure}
    \hspace{-0.6em}
    \begin{subfigure}{0.24\textwidth}
        \centering
        \includegraphics[height=3.54cm]{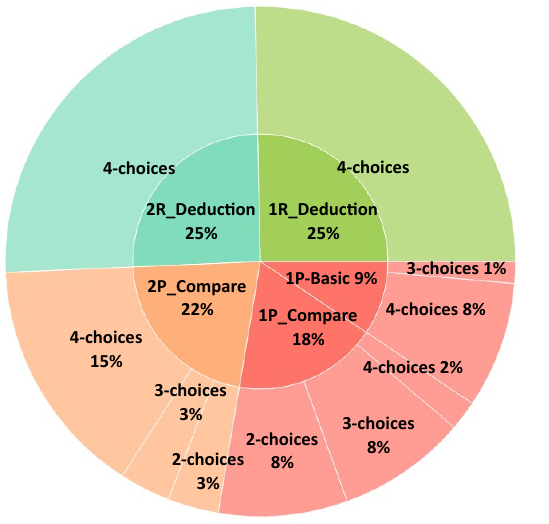}
        \caption{Num. Choice Distribution}
        \label{fig:analysis_3}
    \end{subfigure}
    \hspace{-0.2em}
    \begin{subfigure}{0.24\textwidth}
        \centering
        \includegraphics[height=3.54cm]{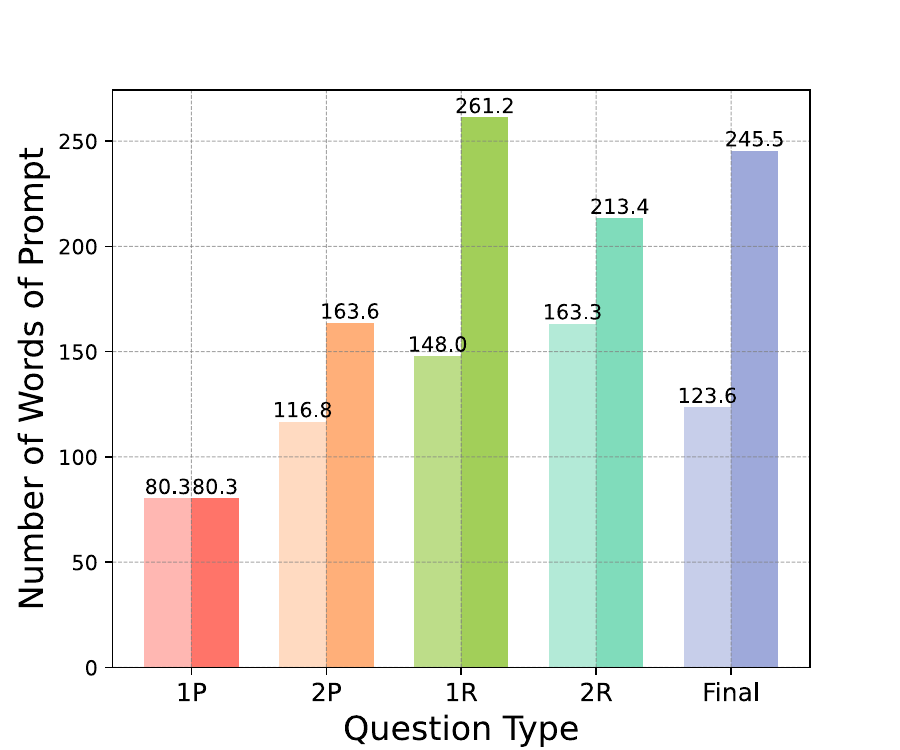}
        \caption{Prompt Length}
        \label{fig:analysis_4}
    \end{subfigure}
    \caption{The left three panels (a), (b), and (c) present analyses of the Direct Answer task, while the right panel (d) focuses on the Logical Chain task.}
    \label{fig:dataset_analysis}
\end{figure*}
\appendix
\section{Appendix}
\label{sec:appendix}
\subsection{Dataset Analysis}
\label{sec:Dataset Analysis}

Figure \ref{fig:analysis_1} compares question lengths across different reasoning VQA datasets. Our dataset stands out with a roughly even distribution of question lengths, unlike other datasets that predominantly focus on shorter questions, which makes our dataset more challenging for MLLMs. Figure \ref{fig:analysis_2} illustrates the proportion of functional programs used in the dataset, showing a wide variety of functions, with query\_rule being slightly more frequent. Figure \ref{fig:analysis_3} highlights the number of multiple-choice options for each configuration, where differences in the number of choices arise due to constraints in the answer space for some configurations. Figure \ref{fig:analysis_4} presents the input prompt length for each stage in the logical chain task, comparing settings with and without prior information. Incorporating prior information from earlier stages significantly increases the maximum prompt length to 261.2 tokens, posing a challenge for MLLMs to parse effectively. 

\begin{table*}[ht]
\begin{center}
    \centering
    \begin{adjustbox}{max width=\textwidth}
    \renewcommand{\arraystretch}{1.3}
    \begin{tabular}{p{5cm}p{5cm}p{1.5cm}p{3cm}p{4cm}}
    \hline
    \textbf{Question Pattern} &
    \textbf{Question Example} &
    \textbf{Attribute} &
    \textbf{Constraints} &
    \textbf{Answer Space} \\
    \hline
    \multicolumn{5}{c}{\textbf{One-Panel Basic}}\\
    \hline
        How many objects are in panel <P>? & How many objects are in panel 1? & Number & NA & [1,2,3,4,5,6,7,8,9]  \\ 
        What is the shape of the object at <X> in panel <P>? & What is the shape of the object at top-left in panel 1? & Shape & NA & [
      "triangle",
      "square",
      "pentagon",
      "hexagon",
      "circle"
    ]  \\ 
        Where is the <S> positioned in panel <P>? & Where is the triangle positioned in panel 1? & Position & NA & ["Left", "Right", "Top", "Down", "Bottom-Left", ...] \\ \hline
    \multicolumn{5}{c}{\textbf{One-Panel Comparison}}\\
    \hline
        In panel <P>, is the shape of the object on the <X> have the same, more, or fewer edges compared to the object on the <X2>? (Note: The edge number increases in the following order: triangle, square, pentagon, hexagon, circle) & In this panel, is the shape of the object on the left have the same, more, or fewer edges compared to the object on the right? (Note: The edge number increases in the following order: triangle, square, pentagon, hexagon, circle)" & Shape & Not\_Equal(X, X2) & [
      "The same",
      "Fewer",
      "More"
    ]  \\ 
        In panel <P>, does the object on the <X> the same, smaller or larger in size compared to the object on the <X2>? & In panel 1, does the object on the top-left the same, smaller or larger in size compared to the object on the bottom-right? & Size & Not\_Equal(X, X2) & [
      "The same",
      "Smaller",
      "Larger"
    ]  \\ 
        In panel <P>, does the object on the <X> the same, darker or brighter in color compared to the object on the <X2>?  & In panel 1, does the object on the top-left the same, darker or brighter in color compared to the object on the bottom-right? & color & Not\_Equal(X, X2) & [
      "The same",
      "Darker",
      "Brighter"
    ] \\
        In panel <P>, where is the <S> relative to the <S2>? & In panel 1, where is the triangle relative to the square? & Position & Not\_Equal(S, S2) & [
      "Left",
      "Right",
      "Above",
      "Below", ...
    ] \\
        Are all objects in panel <P> of the same shape? & Are all objects in panel 1 of the same shape? & Shape & NA & [
      "Yes",
      "No"
    ] \\
        Are all objects in panel <P> of the same size? & Are all objects in panel 1 of the same size? & Size & NA & [
      "Yes",
      "No"
    ] \\
        Are all objects in panel <P> of the same color? & Are all objects in panel 1 of the same color? & Color & NA & [
      "Yes",
      "No"
    ] \\\hline
    \multicolumn{5}{c}{\textbf{Two-Panels Comparison}}\\
    \hline
        Does panel <P> contain the same number of objects, more objects, or fewer objects than panel <P2>? & Does panel 1 contain the same number of objects, more objects, or fewer objects than panel 2? & Number & Not\_Equal(P, P2), Same\_Row(P, P2) & [
      "The same",
      "More",
      "Fewer"
    ]  \\ 
        Is the shape of all the objects in panel <P> have the same, more, or fewer edges compared to the objects in panel <P2>? If the shapes within either panel are already different from each other, select 'Not Comparable.' (Note: The edge number increases in the following order: triangle, square, pentagon, hexagon, circle) & Is the shape of all the objects in panel 1 have the same, more, or fewer edges compared to the objects in panel 2? If the shapes within either panel are already different from each other, select 'Not Comparable.' (Note: The edge number increases in the following order: triangle, square, pentagon, hexagon, circle) & Shape & Not\_Equal(P, P2), Same\_Row(P, P2) & [
      "The same",
      "Fewer",
      "More",
      "Not comparable"
    ]  \\ 
        Is the size of all the objects in panel <P> the same as, smaller or larger than the objects in panel <P2>? If the sizes within either panel are already different from each other, select 'Not Comparable.' & Is the size of all the objects in panel 1 the same as, smaller or larger than the objects in panel 2? If the sizes within either panel are already different from each other, select 'Not Comparable.' & Size & Not\_Equal(P, P2), Same\_Row(P, P2) & [
      "The same",
      "Smaller",
      "Larger",
      "Not comparable"
    ] \\ \hline
    \end{tabular}
    \end{adjustbox}
        \end{center}
\end{table*}

\begin{table*}[ht]
\begin{center}
    \centering
    \begin{adjustbox}{max width=\textwidth}
    \renewcommand{\arraystretch}{1.3}
    \begin{tabular}{p{6cm}p{5cm}p{1.5cm}p{3cm}p{4cm}}
    \hline
    \textbf{Question Pattern} &
    \textbf{Question Example} &
    \textbf{Attribute} &
    \textbf{Constraints} &
    \textbf{Answer Space} \\
    \hline
    \multicolumn{5}{c}{\textbf{Two-Panels Comparison}}\\
    \hline
        Is the color of all the objects in panel <P> the same as, darker or brighter than the objects in panel <P2>? If the colors within either panel are already different from each other, select 'Not Comparable.' & Is the color of all the objects in panel 1 the same as, darker or brighter than the objects in panel 2? If the colors within either panel are already different from each other, select 'Not Comparable.' & Color & Not\_Equal(P, P2), Same\_Row(P, P2) & [
      "The same",
      "Darker",
      "Brighter",
      "Not comparable"
    ] \\
        Is the position of all the objects in panel <P> the same as the objects in panel <P2>? & Is the position of all the objects in panel 1 the same as the objects in panel 2? & Position & Not\_Equal(P, P2), Same\_Row(P, P2) & [
      "Yes",
      "No"
    ] \\ \hline
    \multicolumn{5}{c}{\textbf{One-Row Rule Deduction}}\\
    \hline
        Examine the three panels in the image from left to right and identify the rule that governs the number of the objects. & - & Number & NA & See Number Rule in Table \ref{table:placeholders}  \\ 
        Examine the three panels in the image from left to right and identify the rule that governs the position of the objects. & - & Position & NA & See Position Rule in Table \ref{table:placeholders}  \\ 
        Examine the three panels in the image from left to right and identify the rule that governs the shape of the objects. & - & Shape & NA & See Shape Rule in Table \ref{table:placeholders}  \\
        Examine the three panels in the image from left to right and identify the rule that governs the size of the objects. & - & Size & NA & See Size Rule in Table \ref{table:placeholders}  \\
       Examine the three panels in the image from left to right and identify the rule that governs the color of the objects. & - & Color & NA & See Color Rule in Table \ref{table:placeholders} \\ \hline
    \multicolumn{5}{c}{\textbf{Two-Rows Rule Deduction}}\\
    \hline
        Inspect the first row of three panels from left to right and inspect the second row of three panels from left to right and determine a rule applicable to both rows that governs the number of objects. & - & Number & NA & Same as One-Row  \\ 
        Inspect the first row of three panels from left to right and inspect the second row of three panels from left to right and determine a rule applicable to both rows that governs the position of objects. & - & Position & NA & Same as One-Row  \\ 
        Inspect the first row of three panels from left to right and inspect the second row of three panels from left to right and determine a rule applicable to both rows that governs the shape of objects. & - & Shape & NA & Same as One-Row  \\
        Inspect the first row of three panels from left to right and inspect the second row of three panels from left to right and determine a rule applicable to both rows that governs the size of objects. & - & Size & NA & Same as One-Row  \\
        Inspect the first row of three panels from left to right and inspect the second row of three panels from left to right and determine a rule applicable to both rows that governs the color of objects. & - & Color & NA & Same as One-Row \\ \hline
    \end{tabular}
    \end{adjustbox}
        \end{center}
    \caption{Question pattern templates with corresponding example questions. There are 25 templates in total. 3 templates for One-Panel Basic. 7 templates for One-Panel Comparison. 5 templates for Two-Panels Comparison. 5 Templates for One-Row Rule Deduction. 5 Templates for Two-Rows Rule Deduction. }
    \label{tab:templates}
\end{table*}

\begin{table*}[ht]
\renewcommand{\arraystretch}{1.5}
\begin{tabularx}{\textwidth}{lX}
\hline
\textbf{Category} & \textbf{Value Ranges} \\
\hline
\multicolumn{2}{c}{\textbf{Placeholders}}\\
    \hline
Position (<X>) &
center, left, right, top, bottom, top-left, top-right, bottom-left, bottom-right, top-left, top-center, top-right, middle-left, middle-center, middle-right, bottom-left, bottom-center, bottom-right, outer-part, inner-part, top-left of the inner part, top-right of the inner part, bottom-left of the inner part, bottom-right of the inner part \\
\hline
Panel (<P>) &
0, 1, 2, 3, 4, 5, 6, 7 \\
\hline
Shape (<S>) &
triangle, square, pentagon, hexagon, circle \\
\hline
\multicolumn{2}{c}{\textbf{Rules}}\\
    \hline
Number Rule &
The number of objects gradually decreases by 1; The number of objects remains constant; The number of objects gradually increases by 1; The number of objects distributes three distinct values across panels, rotating through each possible permutation of these values; The number of objects in the last panel equals the sum of the objects in the previous two panels; The number of objects in the last panel equals the difference between the objects in the previous two panels; No clear rule is present \\
\hline
Position Rule &
If an object is in the first panel but not in the second at corresponding position, it appears in the third panel; The position of objects in the last panel is the union of positions from the previous two panels; Three distinct position settings across panels, rotating through each possible permutation of these settings; The position of objects does not change across panels; No clear rule is present \\
\hline
Color Rule &
The color of objects gradually darkens by a constant amount each time; The color of objects gradually brightens by a constant amount each time; The color of objects in the last panel is the sum of the colors in the previous two panels; The color of objects in the last panel is the difference between the colors in the previous two panels; Three distinct colors across panels, rotating through each possible permutation of these colors; The color remains constant; No clear rule is present \\
\hline
Size Rule &
The size of objects gradually increases by a constant amount each time; The size of objects gradually decreases by a constant amount each time; The size of objects in the last panel is the sum of the sizes in the previous two panels; The size of objects in the last panel is the difference between the sizes in the previous two panels; Three distinct sizes across panels, rotating through each possible permutation of these sizes; The size remains constant; No clear rule is present \\
\hline
Shape Rule &
The edge number of shape gradually decreases by 1; The edge number of shape gradually increases by 1; Three distinct shapes across panels, rotating through each possible permutation of these shapes; The shape remains constant; No clear rule is present \\
\hline
\end{tabularx}
\caption{Pre-defined placeholder value ranges and rules for five attributes}
\label{table:placeholders}
\end{table*}

\subsection{Generation Template}
\label{sec:Generation Template}
Table \ref{tab:templates} outlines the question templates used for the Direct Answer Task. Notably, it is impractical to ask questions about the color or size of a single object, as these attributes are represented by numerical values (e.g., color as 255 or size as 8), which models cannot interpret meaningfully. Therefore, questions about size and color are excluded from the basic one-panel tasks. Instead, comparative questions such as "darker" or "smaller" are included in the one-panel comparison tasks. Two constraints are applied during template creation: Not\_Equal(P, P2) ensures that Panels P and P2 are different, and Same\_Row(P, P2) ensures that Panels P and P2 belong to the same row. For answer spaces exceeding four options, a sampling method is used to limit the choices to a maximum of four. Table \ref{table:placeholders} lists all possible values that each placeholder can take, as well as the complete set of rules for each attribute in the rule deduction configuration. After the placeholder values are assigned, references to "Panel <P>" are replaced with "this panel" to enhance clarity and readability.

\subsection{The Full Logical Chain}
\label{sec:The Full Logical Chain}
\begin{figure}[ht]
    \centering
    \includegraphics[width=0.49\textwidth]{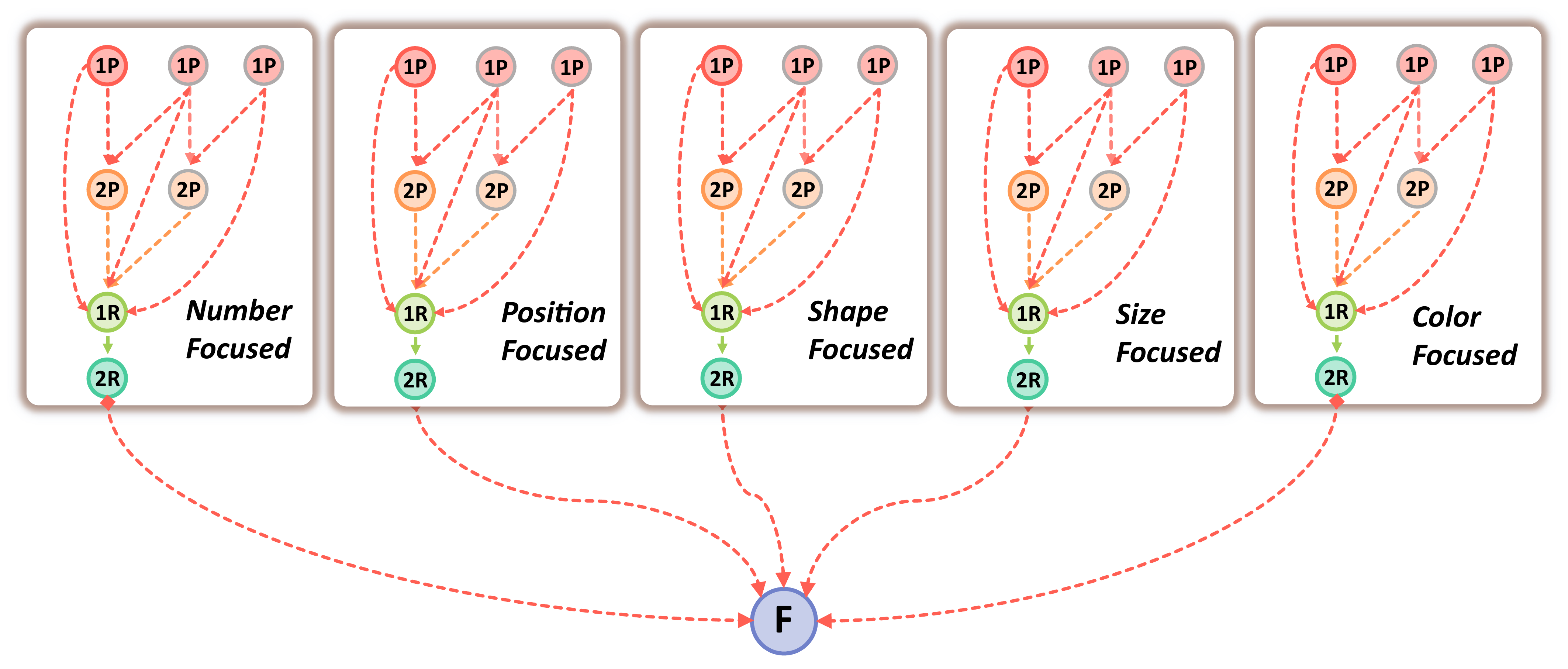}
    \caption{The full logical chain: To arrive at the final answer, we incorporate rules from all five attributes.}
    \label{fig:full_chain}
\end{figure}

Figure \ref{fig:full_chain} presents the full view of our pre-defined logical chain, while Table \ref{tab:logical_chain_examples} provides corresponding question examples for each node in the chain. The chain's ultimate objective is to solve the original RAVEN puzzle, where each row contains rules for attributes such as number, position, shape, size, and color. Intuitively, we connect each attribute's rule deduction phase to the final phase, operating under the assumption that knowing all hidden rules of the RAVEN puzzle provides sufficient information to solve it.

To determine the rules, we expand the reasoning scope from one row to two rows. For one-row rule deduction, we link single-panel perception and two-panel comparison to ensure that with panel-level details and inter-panel comparisons, the rule can be identified. This logical chain is crafted to mimic human problem-solving behavior: focusing first on single-panel perception, followed by panel comparisons, then deducing the first-row rule and validating it with the second row.

However, this handcrafted chain design has limitations. First, it may not align with the model's actual reasoning process, which can cause discrepancies in performance. Additionally, to simplify chain construction, we designed it at the "Corpus-Level," meaning it remains fixed across all instances. This approach sometimes results in insufficient prior information for certain stages. For example, in one-row rule deduction, a rule like "the number of objects distributes three distinct values across panels, rotating through each possible permutation" may require second-row information to resolve. These limitations highlight the need for more flexible and instance-specific logical chain designs in future work.

\begin{table*}[h!]
\centering
\renewcommand{\arraystretch}{1.1}
\small
\resizebox{0.98\textwidth}{!}{
\begin{tabular}{|>{\centering\arraybackslash}m{0.15\textwidth}|m{0.7\textwidth}|}
\hline
\textbf{Attributes} & \textbf{Logical Chain Stage Example} \\
\hline
\textbf{Number} & \textcolor{lowcontrastred}{1P}: How many objects are in the panel? \newline
\textcolor{lowcontrastyellow}{2P}: Does the left panel contain the same number of objects, more objects, or fewer objects than the right panel? \newline
\textcolor{lowcontrastgreen}{1R}: Inspect the three panels in the image from left to right and identify the rule that dictates the number of objects. \newline
\textcolor{lowcontrastblue}{2R}: Inspect the first row of three panels from left to right and inspect the second row of three panels from left to right and determine a rule applicable to both rows that governs the number of objects. \\
\hline
\textbf{Position} & \textcolor{lowcontrastred}{1P}: Where is the circle positioned in the panel? \newline
\textcolor{lowcontrastyellow}{2P}: Is the position of all the objects in the left panel the same as the objects in the right panel? \newline
\textcolor{lowcontrastgreen}{1R}: Examine the three panels in the image from left to right and identify the rule that governs the position of the objects. \newline
\textcolor{lowcontrastblue}{2R}: Examine the three panels in the first row, then the three panels in the second row, both from left to right, and derive a rule that applies to both rows in relation to the position of objects. \\
\hline
\textbf{Shape} & \textcolor{lowcontrastred}{1P}: What is the shape of the object at center in the panel? \newline
\textcolor{lowcontrastyellow}{2P}: Is the shape of all the objects in the left panel have the same, more, or fewer edges compared to the objects in the right panel? \newline
\textcolor{lowcontrastgreen}{1R}: Inspect the three panels in the image from left to right and identify the rule that dictates the shape of objects. \newline
\textcolor{lowcontrastblue}{2R}: Analyze the first row of three panels from left to right, followed by the second row of three panels, and identify a common rule that dictates the shape of objects in both rows. \\
\hline
\textbf{Size} & \textcolor{lowcontrastred}{1P}: Are all objects in the panel of the same size? \newline
\textcolor{lowcontrastyellow}{2P}: Is the size of all the objects in the left panel the same as, smaller or larger than the objects in the right panel? \newline
\textcolor{lowcontrastgreen}{1R}: Analyze the three panels in the image from left to right and uncover the rule that governs the size of objects. \newline
\textcolor{lowcontrastblue}{2R}: Review the first row of three panels in sequence from left to right, then do the same for the second row, and determine a shared rule that governs the size of objects in both rows. \\
\hline
\textbf{Color} & \textcolor{lowcontrastred}{1P}: Are all objects in the panel of the same color? \newline
\textcolor{lowcontrastyellow}{2P}: Is the color of all the objects in panel $<$P$>$ the same as, darker or brighter than the objects in panel $<$P2$>$? \newline
\textcolor{lowcontrastgreen}{1R}: Inspect the three panels in the image from left to right and identify the rule that dictates the color of objects. \newline
\textcolor{lowcontrastblue}{2R}: Examine the three panels in the first row, then the three panels in the second row, both from left to right, and derive a rule that applies to both rows in relation to the color of objects. \\
\hline
\textbf{Final} & You are presented with a 3x3 grid of panels, called the \textit{Problem Matrix.} The last panel is missing and marked with a `?' symbol. Below the matrix, there is a set of 8 possible answer options labeled from 1 to 8. Your task is to determine which panel from the answer set (1-8) correctly fits the missing position in the problem matrix. The pattern in the matrix follows some hidden rules that apply row by row (horizontally). Please select the number (from 1 to 8) of the panel that completes the pattern. \\
\hline
\end{tabular}
}
\caption{A full logical chain with the examples for five stages.}
\label{tab:logical_chain_examples}
\end{table*}

\subsection{Input Prompt And Model Settings}
\label{sec:Input Prompt to MLLMs}
\subsubsection{Direct Answer}
\label{sec:Direct Answer}
\paragraph{Model Settings:}All MLLMs are tested under their default settings under the environment of Huggingface \footnote{\url{https://huggingface.co/}}, the \textit{transformer} package version in python is 4.39.2 for NVLM-D-72B model and 4.46.0 for all others. 

\begin{figure}[ht]
  \centering
   \includegraphics[width=\linewidth]{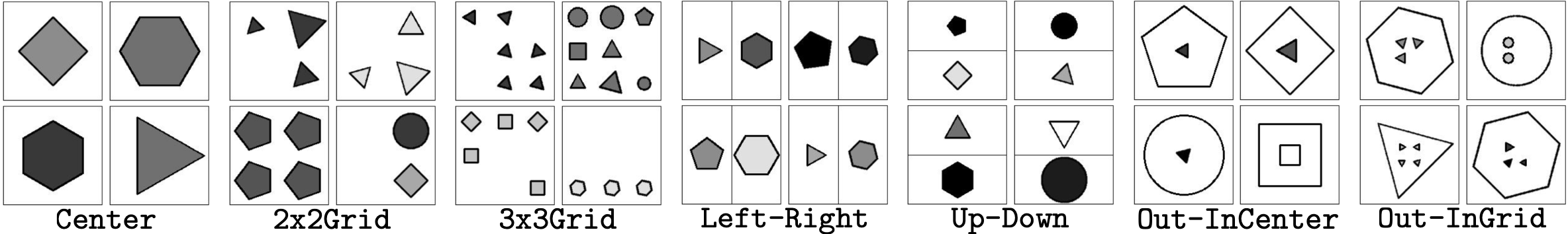}
   \caption{The original RAVEN puzzle, includes seven puzzle settings.}
   \label{fig:RAVEN}
\end{figure}

\paragraph{Prompt Details:}The RAVEN dataset includes various puzzle settings, such as Left-Right, Up-Down, and In-Out, where rules are applied separately to distinct parts of the panels (Figure \ref{fig:RAVEN}). To address these settings, when we decompose the problem into subproblems, we treat each part independently. For instance, there are separate question sets for the left and right sections of the panels in the Left-Right setting, with the question explicitly stating which part is being addressed. To clarify the panel structure, an additional sentence is appended to the question:

\begin{itemize}
    \item \textbf{Left-Right:} The panel is divided into two sections by a vertical line, separating the left side from the right side, with objects possibly present in both sections.
    \item \textbf{Up-Down:} The input panel is split by a horizontal line, separating the top side from the bottom side, with objects possibly present in both sections.
    \item \textbf{In-Out:} The panel is divided into two regions: an outer structure and an inner structure, with objects possibly present in both regions.
\end{itemize}

This extra information is unnecessary for other settings. The complete prompt format is:
\textbf{[}Extra Setting Info\textbf{] Question: }[question]\textbf{ Please select one of the following: }[choices]\textbf{. The answer should be one of A, B, C, D.}

\subsubsection{Logical Chain}
\label{sec:Logical Chain}
\paragraph{Model Settings:}All MLLMs are tested under their default settings under the environment of Huggingface \footnote{\url{https://huggingface.co/}}, the \textit{transformer} package version in python is 4.39.2 for NVLM-D-72B model and 4.46.0 for all others. In addition, to handle the length of our prompts, we increase the maximum token length to 2048.

When prior information is injected, a rule-based program is used to convert the information into text and integrate it into the prompt. This transformation is necessary because the images referenced in prior questions are not the same as those in the current question, making it impossible to directly reuse them.

For example, if the prior question is "How many objects in this panel?" and the current question is "Comparing the number of objects in the left and right panel," the phrase "this panel" cannot directly correspond to "left panel" or "right panel." To address this, we transform "this panel" into a more specific term, such as "left panel" or "right panel." Table \ref{tab:rule_based_program} outlines the transfer rules for Number and Position. Similar patterns are applied for other attributes, which are not listed here for brevity.

After the prior information is transformed, the prompt is structured as follows:
\textbf{[}Extra Setting Info\textbf{] Below is the information generated from the previous steps, please be aware that it may or may not contain errors: }[[Prior Info 1], [Prior Info 2], ...]\textbf{ Question: }[question]\textbf{ Please select one of the following: }[choices]\textbf{. The answer should be one of A, B, C, D.}

\begin{table}[ht]
\setlength{\tabcolsep}{4pt}
\renewcommand{\arraystretch}{1.1}
\small
\resizebox{0.50\textwidth}{!}{
\begin{tabular}{lccccccc}
\hlineB{4}  
                                & Metric                      & Prior & 1P     & 2P     & 1R     & 2R     & Final       \\ \hline\hline
\multirow{4}{*}{\makecell{InstructBLIP}}    & \multirow{2}{*}{Acc}   & w/o        & 31.94   & 28.09   & 22.00   & 23.82   & 7.14        \\
                                &                            & w          & 31.94   & 28.09   & 22.13   & 23.45   & 8.57        \\ \cline{2-8}
                                & \multirow{2}{*}{MSEval}     & w/o        & 1.000   & 1.000   & 1.007   & 1.000   & 0.848        \\
                                &                            & w          & 1.000   & 1.000   & 1.004   & 1.004   & 0.959        \\ \cline{1-8}
\multirow{4}{*}{xGen-MM}   & \multirow{2}{*}{Acc}   & w/o        & 60.48   & 23.27   & 25.82   & 21.64   & 14.29       \\
                                &                            & w          & 60.48   & 23.18   & 20.18   & 26.36   & 5.71 \\ \cline{2-8}
                                & \multirow{2}{*}{MSEval}     & w/o        & 2.006   & 0.794   & 0.965   & 0.953   & 1.290 \\
                                &                            & w          & 2.006   & 1.082   & 0.812   & 0.985   & 1.013 \\ \cline{1-8}
\multirow{4}{*}{Llava-13b} & \multirow{2}{*}{Acc}   & w/o        & 30.91   & 31.55   & 23.09   & 22.36   & 15.71        \\
                                &                            & w          & 30.91   & 31.09   & 22.36   & 21.64   & 15.71        \\ \cline{2-8}
                                & \multirow{2}{*}{MSEval}     & w/o        & 1.059   & 0.997   & 0.941   & 0.910   & 1.002        \\
                                &                            & w          & 1.059   & 1.009   & 0.944   & 0.929   & 0.975        \\ \cline{1-8}
\multirow{2}{*}{Random}         & Acc   & -          & 31.1   & 31.7   & 25.0   & 25.0   & 12.5        \\
                                & MSEval     & -          & 1.00   & 1.00   & 1.00   & 1.00   & 1.00        \\ \hlineB{4}
\end{tabular}
}
\caption{The Accuracy (Acc) and MSEval scores for the Logical Chain task. w/o: without prior, w: with prior.}
\label{table:more results}
\end{table}

\paragraph{More results:} Table \ref{table:more results} shows the results three additional models InstructBlip-13B, LLaVA-1.5-13B, and xgen-mm. Prior information appears to provide limited utility for these models, all of them are just around random baselines except xgen-mm have relative good basic perception ability.

\begin{table*}[ht]
\centering
\begin{tabular}{@{}lllp{9cm}@{}}
\toprule
\textbf{Attribute} & \textbf{Stage}       & \textbf{Numbers}        & \textbf{Output} \\ \midrule
\textbf{Number}    & \texttt{single\_panel}   & \texttt{["1"]}           & There are \texttt{\{answer\_str\}} objects in the left panel. \\ 
                   &                       & \texttt{["2"]}           & There are \texttt{\{answer\_str\}} objects in the right panel. \\ 
                   &                       & \texttt{["3"]}           & There are \texttt{\{answer\_str\}} objects in the right panel. \\ 
                   & \texttt{two\_panels}     & \texttt{["1", "2"]}      & The left panel has \texttt{\{answer\_str\}} objects compared to the middle panel. \\ 
                   &                       & \texttt{["2", "3"]}      & The middle panel has \texttt{\{answer\_str\}} objects compared to the right panel. \\ 
                   & \texttt{one\_row}        & Any                     & The rule for the number of objects in the first row is: \texttt{\{answer\_str\}}. \\ \midrule
\textbf{Position}  & \texttt{single\_panel}   & \texttt{["1"]}           & \texttt{Where is the (\textbackslash w+) positioned in the panel?} becomes: \texttt{There is a \textbackslash 1 positioned in the left panel.} \\ 
                   &                       & \texttt{["2"], ["3"]}    & \texttt{Where is the (\textbackslash w+) positioned in the panel?} becomes: \texttt{There is a \textbackslash 1 positioned in the right panel.} \\ 
                   & \texttt{two\_panels}     & \texttt{["1", "2"]}      & If \texttt{answer\_str} is \texttt{Yes}, "The position of all the objects in the left panel is the same as the objects in the middle panel." Otherwise, "The position of all the objects in the left panel is not the same as the objects in the middle panel." \\ 
                   &                       & \texttt{["2", "3"]}      & If \texttt{answer\_str} is \texttt{Yes}, "The position of all the objects in the middle panel is the same as the objects in the right panel." Otherwise, "The position of all the objects in the middle panel is not the same as the objects in the right panel." \\ 
                   & \texttt{one\_row}        & Any                     & The rule for the position of objects in the first row is: \texttt{\{answer\_str\}}. \\ \bottomrule
\end{tabular}
\caption{Rule-based program of attribute Number and Position. The Stage represents the prior stage. (\textbackslash w+) represents the word here will be put in the position of \textbackslash 1.}
\label{tab:rule_based_program}
\end{table*}

\label{sec:error analysis}
\subsection{Error Analysis}
\begin{figure}[ht]
    \centering
    \includegraphics[width=0.47\textwidth]{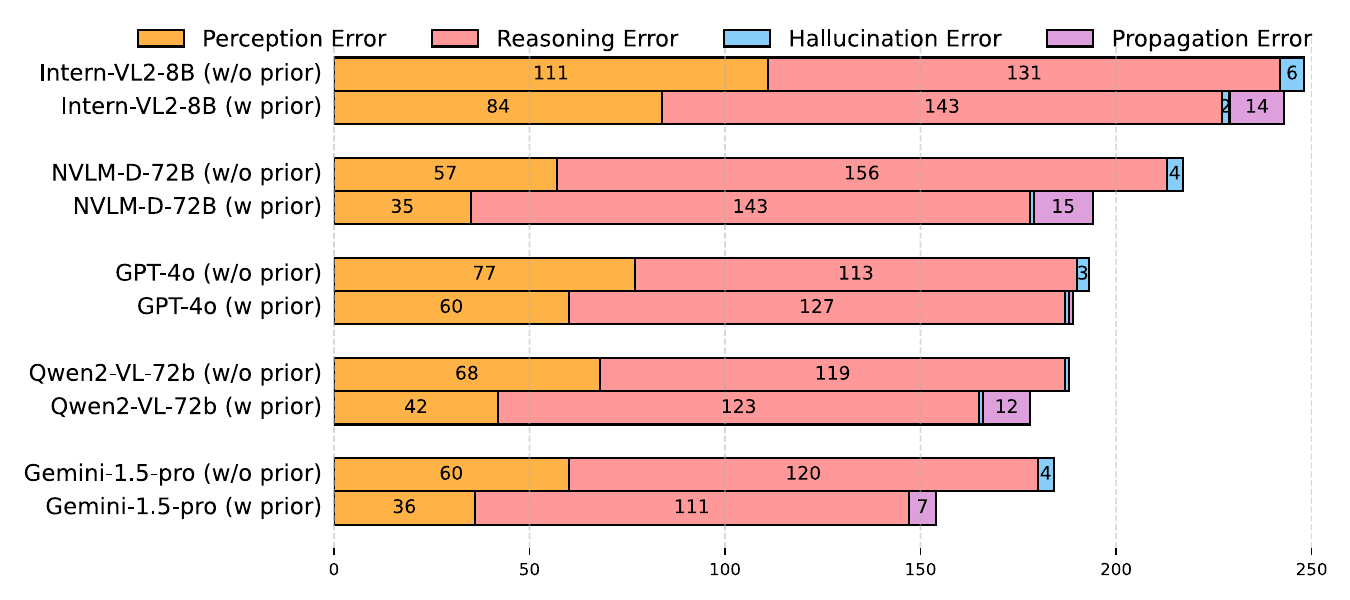}
    \caption{Errors distribution for each model under the settings of without and with prior.}
    \label{fig:error_analysis}
\end{figure}

\subsubsection{Errors from Explanation}
To further investigate model performance, we conduct an error analysis for the Logical Chain subtask. Models are asked to generate with explanations alongside answers, and we manually reviewed all output explanations when the models predict incorrect answer. Errors are classified into four types: 
\begin{itemize}
    \item \textbf{Perception Error:} This occurs when the model misinterprets visual inputs, such as object numbers or shapes. In the provided example, there should be four objects in the left panel and three in the right panel, but the model fails to recognize this correctly.

    \item \textbf{Reasoning Error:} This involves incorrect logic applied to correctly perceived inputs. In this example, the model accurately identifies the objects and their edge numbers in each panel. However, due to flawed reasoning, it incorrectly concludes that the number of edges is decreasing, which is not the case.

    \item \textbf{Unrelated Information Error:} This error type refers to the generation of incomplete sentences or unrelated information. Here, the question asks about the position of objects, but the explanation provided by the model focuses on shapes and sizes, which are irrelevant.

    \item \textbf{Propagation Error:} This occurs when the model fails to detect or correct inaccuracies in prior information. In this example, the prior information is already incorrect, but the model does not identify or address these inaccuracies, leading to an incorrect answer.
\end{itemize}

Figure \ref{fig:error_analysis} reveals that reasoning errors are the most common, followed by perception errors, with unrelated-information and propagation errors being rare. Gemini exhibits the lowest perception error rate, while GPT-4o shows the lowest reasoning error rate. Notably, injecting prior information significantly reduces perception errors, demonstrating that prior knowledge enhances models' understanding of visual inputs, but it does not help with reasoning error. 

\subsubsection{Unanswerable Question}
We conducted a small experiment by making all the answer choices incorrect, this can be used to verify the models robustness. In this experiment, we defined hallucination as the model's failure to recognize that the question is "non-answerable". We defined two approaches for setting unanswerable questions:

\begin{itemize}
    \item \textbf{Same Attribute:} All choices are incorrect while retaining the same attribute as the question.
    \item \textbf{Changed Attribute:} The choices are changed to a different attribute (e.g., the question asks about the number of objects, but the choices are their positions).
\end{itemize}

We sampled 10 examples for each stage (60 in total) and tested them on GPT-4o, manually reviewing the output explanations. As shown in Table \ref{table:settings_comparison}, interestingly, as the questions become more difficult, particularly at the final step, the model increasingly fails to distinguish unanswerable questions, resulting in higher rates of hallucination. Furthermore, under Setting 2, where the attribute is changed, the model exhibits a greater likelihood of hallucination, as it struggles to recognize the shift in attributes.

\begin{table}[ht]
\centering
\setlength{\tabcolsep}{6pt}
\renewcommand{\arraystretch}{1.2}
\small
\begin{tabular}{lcccccc}
\hlineB{3}
Setting  & 1P-C  & 1P-B  & 2P   & 1R   & 2R   & Final \\ \hline\hline
Setting 1 & 0/10  & 3/10  & 1/10 & 8/10 & 7/10 & 10/10 \\
Setting 2 & 7/10  & 4/10  & 8/10 & 10/10 & 10/10 & 10/10 \\ \hlineB{3}
\end{tabular}
\caption{Performance comparison of different settings across various stages. Setting 1: Same Attribute; Setting 2: Changed Attribute. The ratio is Hallucinations / Total:}
\label{table:settings_comparison}
\end{table}

\subsection{Human Sudies and and Inter-Participant Agreement}
\label{sec:Human Studies}

To evaluate the subjective quality of human performance in our study, we conducted two separate parts: Part A and Part B. Part A focuses on evaluation of the quality of our automatically generated dataset, while Part B focuses on testing human reasoning abilities across different stages of complexity. For both Part A and Part B, a Consent Form and a Plain Language Statement are provided to the annotators prior to the annotation process. These documents must be read and agreed upon before they can proceed with the annotations.
\subsubsection{Part A}
\label{sec:Part A}
\begin{figure}[ht]
  \centering
   \includegraphics[width=\linewidth]{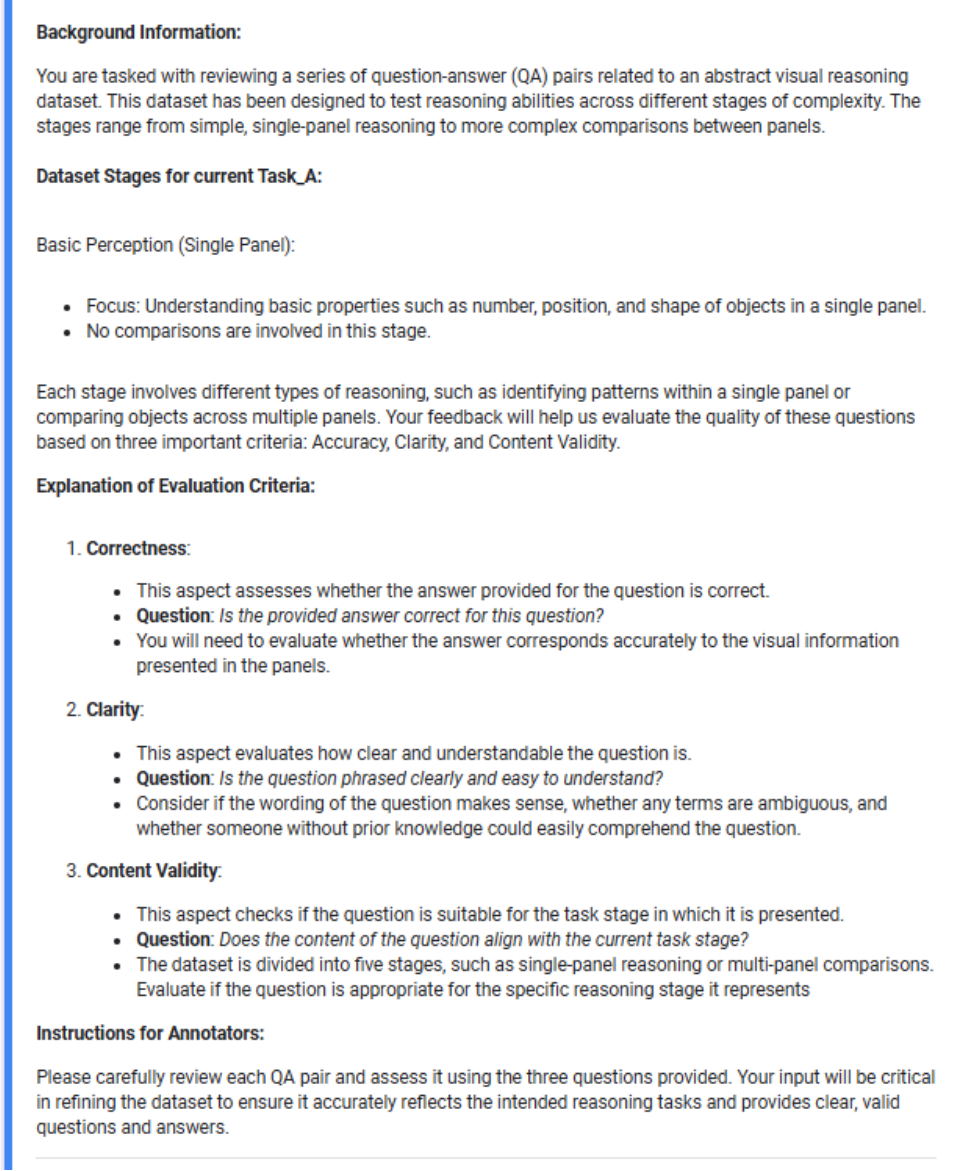}
   \caption{A detailed guide provided to participants at the beginning of the questionnaire for Part A.}
   \label{fig:parta_guideline}
\end{figure}

\begin{figure}[ht]
  \centering
   \includegraphics[width=\linewidth]{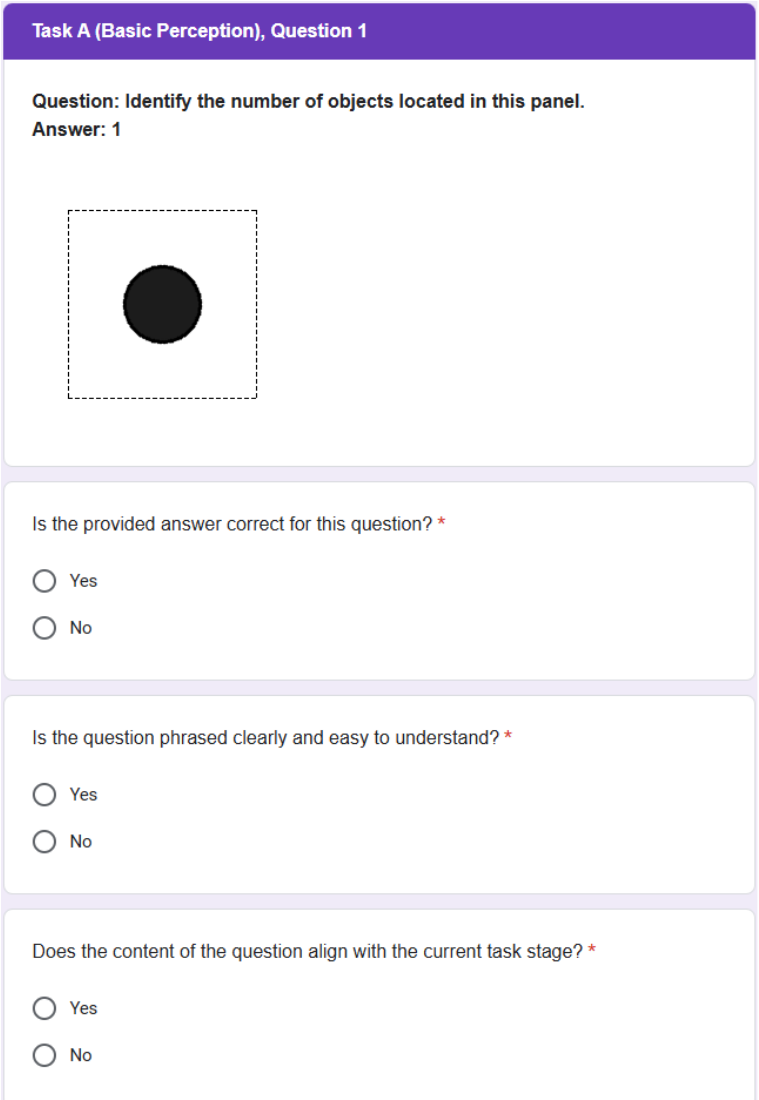}
   \caption{Sample Question from the Questionnaire for Part A.}
   \label{fig:sample_parta}
\end{figure}

Part A involved five research students who participated in answering a series of abstract reasoning questions. This section aimed to evaluate the quality of the dataset generated by our template-based methods. Since the Direct Answer and Logical Chain share the same pool of templates, and Direct Answer covers all templates, we chose to focus on assessing the quality of the Direct Answer component. A random sample of 620 questions was selected for this evaluation. To ensure participants clearly understood the tasks and evaluation criteria, a detailed guide was provided at the beginning of the questionnaire (see Figure \ref{fig:parta_guideline}). An example question from the questionnaire is shown in Figure \ref{fig:sample_parta}.

To thoroughly assess the human performance in Part A, we used three key indicators: Correctness, Clarity, and Content Validity. 

Correctness assesses whether the answer provided for the question is correct. Evaluators were asked to determine if the provided answer accurately corresponded to the visual information presented in the panels. This involved a careful comparison between the answer and the visual data to ensure accuracy. 

Clarity evaluates how clear and understandable the question is. Evaluators considered whether the question was phrased clearly and was easy to understand. They assessed if the wording made sense, if any terms were ambiguous, and whether someone without prior knowledge could easily comprehend the question. This indicator is crucial for ensuring that the questions are accessible and interpretable by all participants. 

Content validity checks if the question is suitable for the task stage in which it is presented. Evaluators examined whether the content of the question aligned with the current task stage. The dataset is divided into five types, such as one panel basic perception or two panel comparisons. Participants needed to ensure that the question was appropriate for the specific reasoning type it represented. This indicator ensures that each question is relevant and appropriately challenging for its designated stage.

The metrics used to evaluate performance in Part A included correctness, clarity, and content validity, with positive rates for each metric provided in Table \ref{tab:parta}. The positive rate is the proportion of questions answered by "Yes". The results indicate that the participants in Part A performed exceptionally well across all metrics, with Correctness, Clarity, and Content Validity scores consistently high. This suggests that the questions were well-designed and comprehensible, and the participants were able to provide accurate answers.

\begin{table*}[ht]
\setlength{\tabcolsep}{4pt}
\renewcommand{\arraystretch}{1}
\small
\centering
\resizebox{\textwidth}{!}{
\begin{tabular}{c|ccccc}
\hline
\textbf{Metric}  & \textbf{One Panel Basic} & \textbf{One Panel Compare} & \textbf{Two Panel Compare} & \textbf{One Row} & \textbf{Two Rows} \\ \hline
Correctness      & 0.98                     & 0.97                       & 0.96                       & 0.93             & 0.94              \\
Clarity          & 0.96                     & 0.97                       & 0.94                       & 0.95             & 0.99              \\
Content Validity & 0.99                     & 0.99                       & 0.99                       & 1.00             & 1.00              \\ \hline
\end{tabular}
}
\caption{Human performance (positive rates) for Part A across different question types.}
\label{tab:parta}
\end{table*}

\subsubsection{Part B}
\label{sec:Part B}
\begin{figure}[ht]
  \centering
   \includegraphics[width=\linewidth]{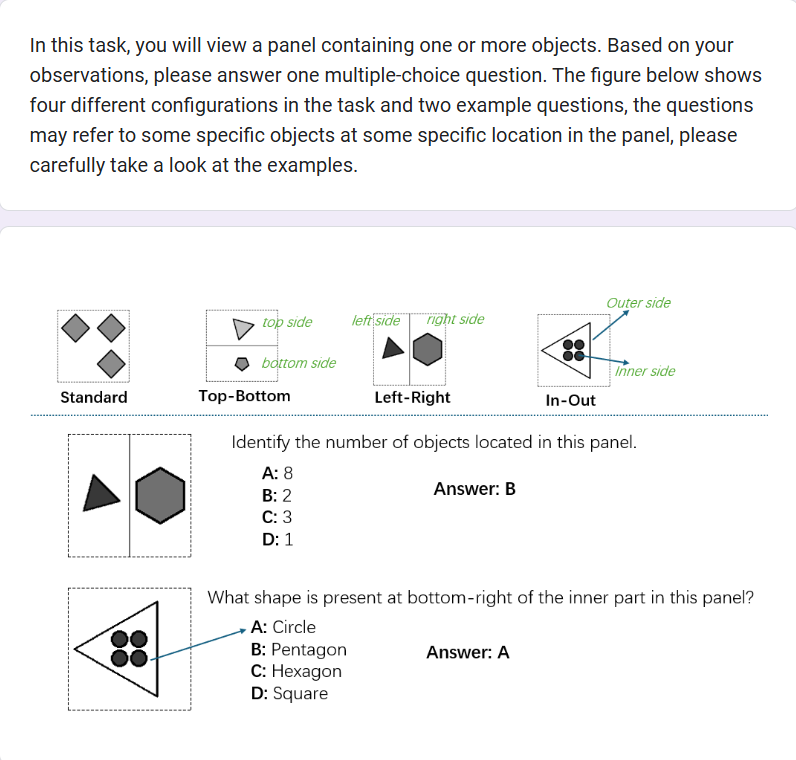}
   \caption{A detailed guide provided to participants at the beginning of the questionnaire for Part B. This guide focuses on One-Panel Basic Perception.}
   \label{fig:partb_guideline}
\end{figure}

\begin{figure}[ht]
  \centering
   \includegraphics[width=\linewidth]{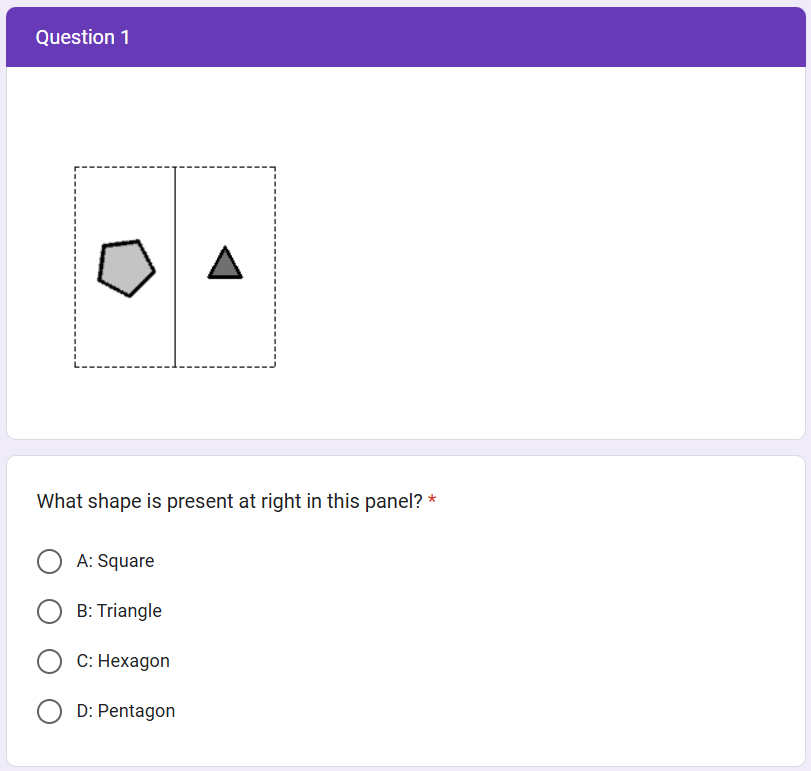}
   \caption{Sample Question from the One-Panel Basic Perception Questionnaire for Part B.}
   \label{fig:sample_partb}
\end{figure}

Part B utilized the Prolific crowdsourcing platform\footnote{\url{https://www.prolific.com/}} to recruit 162 participants who were subjected to the same set of abstract reasoning questions as those given to the MLLMs. The objective of this part was to evaluate human performance on our dataset, enabling a comparison between human and model capabilities. Participants received a detailed guide at the beginning of the questionnaire, which included task descriptions and several examples, as shown in Figure \ref{fig:partb_guideline}. The guide varied depending on the stage of the Direct Answer task, but for this section, we include only the One-Panel Basic Perception stage. And similar to Part A, due to Direct Answer covering all templates, we chose to focus on assessing the human performance of the Direct Answer component. Each question in the questionnaire for Part B included a image and a multiple-choice question, as illustrated in Figure \ref{fig:sample_partb}.

The performance metrics for Part B are summarized in Table \ref{tab:partb}. The performance for Part B show a noticeable decline in positive rates, particularly for more complex tasks such as Two Panel Compare, One Row, and Two Rows. This decline highlights the increased difficulty of these tasks and suggests that the broader participant pool found these questions more challenging.

\begin{table}[ht]
\centering
\resizebox{0.48\textwidth}{!}{
\begin{tabular}{ccccc}
\hline
\textbf{1P-B} & \textbf{1P-C} & \textbf{2P} & \textbf{1R} & \textbf{2R} \\ \hline
98.52           & 88.89             & 69.08             & 62.12   & 63.33    \\ \hline
\end{tabular}
}
\caption{Human performance (positive rates) for Part B across different question types.}
\label{tab:partb}
\end{table}

\textbf{Inter-Participant Agreement.} To quantify the inter-participant agreement across participants for Part B stuides, we computed Fleiss' kappa scores \cite{landis1977measurement} across different question types. The Fleiss' Kappa scores for each question types are provided in Table \ref{tab:kappa}.

\begin{table}[ht]
\centering
\resizebox{0.48\textwidth}{!}{
\begin{tabular}{c|ccccc}
\hline
\textbf{Task} & \textbf{1P-B} & \textbf{1P-C} & \textbf{2P} & \textbf{1R} & \textbf{2R} \\ \hline
Kappa Scores  & 0.9711     & 0.7830     & 0.4988     & 0.4443     & 0.4075     \\ \hline
\end{tabular}
}
\caption{Fleiss' Kappa Scores for Inter-Participant Agreement across different question types.}
\label{tab:kappa}
\end{table}

The high Fleiss' Kappa score for One-Panel Basic (0.9711) indicates strong agreement among the participants, this is mainly due to the simplicity of One-Panel Basic questions. However, the lower scores for Two-Panel and rule deduction phase highlight the increased difficulty and the significant variability in human interpretation for these more complex tasks.

\subsection{Performance Increase with Prior Info}
\label{sec:Performance Increase with Prior Info}
\begin{figure*}[ht] 
    \centering
    \includegraphics[width=\textwidth]{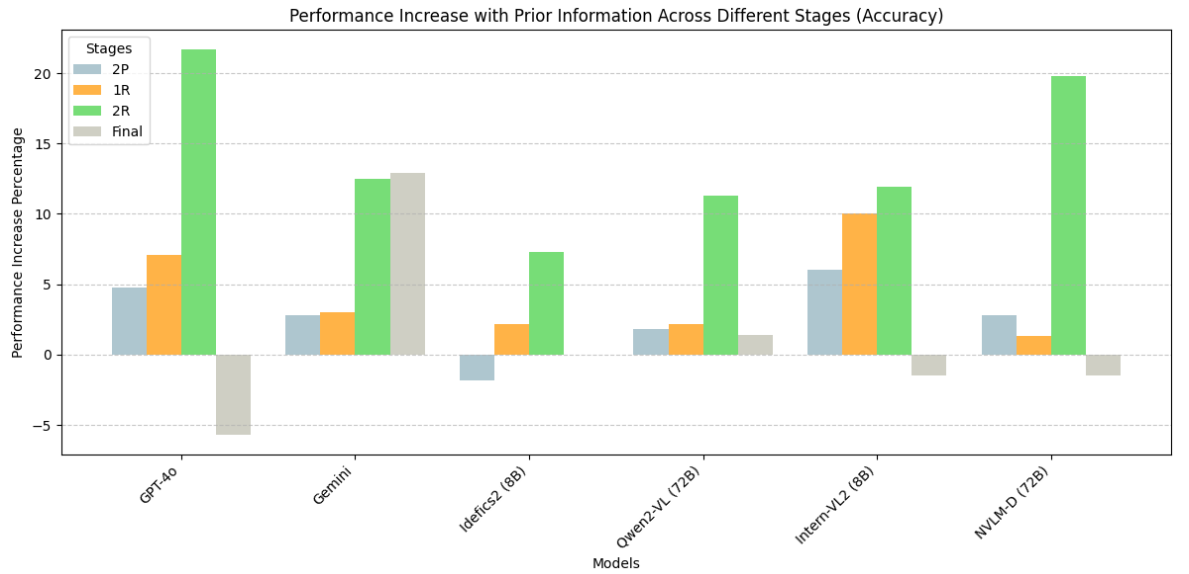} 
    \caption{Accuracy percentage increase after incorporating prior info.}
    \label{fig:increase_accuracy} 
\end{figure*}

\begin{figure*}[ht]
    \centering
    \includegraphics[width=\textwidth]{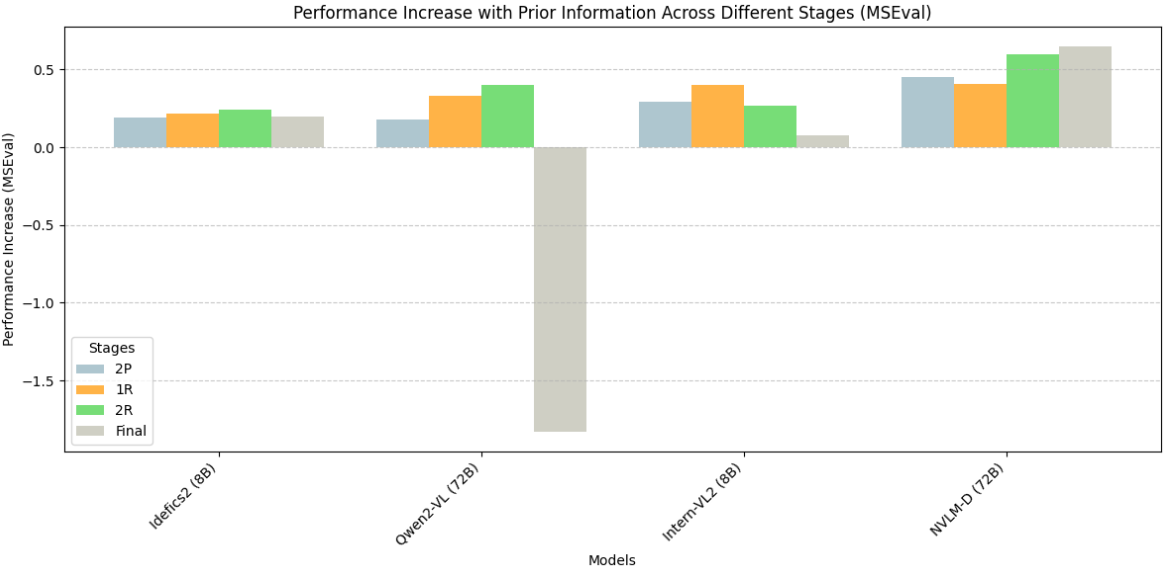} 
    \caption{MSEval percentage increase after incorporating prior info.}
    \label{fig:increase_mseval}
\end{figure*}

Tables \ref{fig:increase_accuracy} and \ref{fig:increase_mseval} present the percentage increase in Accuracy and MSEval metrics, respectively. It is evident that, except for the Final stage, all other stages show improved performance, with MSEval and Accuracy metrics closely aligned in these cases. However, in the Final stage, while Accuracy does not show a significant increase for the four open-source models, MSEval suggests some improvement due to the incorporation of rule information for solving the final RAVEN puzzle. An exception is observed with Qwen2-VL-72B, which may already perform well on RAVEN. Incorporating information from earlier stages might introduce misleading details, leading to a significant performance drop.

\subsection{Additional Details about MSEval}
\label{sec:Additional Details about MSEval}
\subsubsection{Algorithm Pseudo Code}
\label{sec:Algorithm Pseudo Code}
Algorithm \ref{alg:overall_workflow} shows the details Pseudo Code for our proposed MSEval metrics.
\begin{algorithm}[t]
\small
\captionsetup{font=small} 
\caption{Overall Workflow}
\label{alg:overall_workflow}
\begin{algorithmic}
\State \textbf{Input:} 
\State \quad Logical Chain $\mathcal{D}_t$
\State \quad \textbf{Define:} \ensuremath{\mathcal{S}_t = \{t\} \cup \mathcal{D}_t} 
\State \quad Model logits $\mathcal{Z} = \{z_j^{(i)} \mid j \in \mathcal{S}_t, i = 1, \dots, N\}$ 
\State \quad All possible choices for each node $\{\mathcal{A}_j^{(i)} \mid j \in \mathcal{S}_t\}$
\State \textbf{Output:} MSEval score for stage $t$: $\text{MSEval}_t$
\State \textbf{Step 1: Compute Conditional Probabilities}
\For{each sample $i = 1$ to $N$}
    \For{each node $j \in \mathcal{S}_t$}
        \State Compute probability $p_j^{(i)} \gets \frac{\exp(z_j^{(i)})}{\sum_{k \in \mathcal{A}_j^{(i)}} \exp(z_k^{(i)})}$
    \EndFor
\EndFor

\State \textbf{Step 2: Compute Conditional Mutual Information}
\For{each sample $i = 1$ to $N$}
    \For{each node $j \in \mathcal{S}_t$}
        \State Alter $\mathcal{A}_j^{(i)}$ to generate perturbed outputs $\mathcal{A}_{j \to t}^{(i)}$
        \State Compute:
        \[
        \begin{aligned}
        \text{CMI}(i, j, t) \gets 
        &\ H(\mathcal{A}_{j \to t}^{(i)} \mid \mathcal{D}_t^{(i), -j}) \\
        &+ H(\mathcal{A}_j^{(i)} \mid \mathcal{D}_t^{(i), -j}) \\
        &- H(\mathcal{A}_{j \to t}^{(i)}, \mathcal{A}_j^{(i)} \mid \mathcal{D}_t^{(i), -j})
        \end{aligned}
        \]
    \EndFor
\EndFor

\State \textbf{Step 3: Normalize Conditional Mutual Information}
\For{each sample $i = 1$ to $N$}
    \For{each node $j \in \mathcal{S}_t$}
        \State Compute:
        \[
        \text{NCMI}(i, j, t) \gets \frac{\exp(\text{CMI}(i, j, t))}{\sum_{k \in \mathcal{S}_t} \exp(\text{CMI}(i, k, t))}
        \]
    \EndFor
\EndFor

\State \textbf{Step 4: Compute MSEval Score for Each Sample}
\For{each sample $i = 1$ to $N$}
    \State Initialize $\text{MSEval}_t^{(i)} \gets 0$
    \For{each node $j \in \mathcal{S}_t$}
        \State Compute \[ 
        \epsilon_j^{(i)} = \frac{1}{|\mathcal{A}_j^{(i)}|}
        \]
        \State Update:
        \[
        \text{MSEval}_t^{(i)} \gets \text{MSEval}_t^{(i)} + \text{NCMI}(i, j, t) \cdot \frac{p_j^{(i)}}{\epsilon_j^{(i)}}
        \]
    \EndFor
\EndFor

\State \textbf{Step 5: Average MSEval Across All Samples}
\State Compute:
\[
\text{MSEval}_t \gets \frac{1}{N} \sum_{i=1}^N \text{MSEval}_t^{(i)}
\]

\State \textbf{Return:} $\text{MSEval}_t$
\end{algorithmic}
\end{algorithm}

\subsubsection{Computational Cost}
\label{sec:Computational Cost}
The computational complexity of the algorithm can be expressed as:

\[
O(N \cdot |\mathcal{E}_t| \cdot |A|)
\]

where:
\begin{itemize}
    \item \(N\): The number of samples.
    \item \(|\mathcal{E}_t|\): The number of edges in the logical chain (dependency relationships between nodes).
    \item \(|A|\): The number of possible choices for each node.
\end{itemize}

In most cases, \(|A|\), the number of possible choices per node, is typically equal to 4. As a result, the computational complexity simplifies to:
\[
O(4 \cdot N \cdot |\mathcal{E}_t|) \quad \text{or simply} \quad O(N \cdot |\mathcal{E}_t|),
\]
which is effectively linear with respect to both the number of instances (\(N\)) and the number of edges (\(|\mathcal{E}_t|\)) in the logical chain. By reducing the number of edges or employing a smaller logical chain, the computational cost can be significantly minimized, ensuring better scalability and efficiency, especially for large datasets or complex logical dependencies. This simplification highlights the importance of optimizing the chain structure to maintain computational feasibility. The actual time cost for each open-source model we tested is shown in Table \ref{tab:running_time}.

\begin{table}[ht]
\centering
\renewcommand{\arraystretch}{1.2} 
\setlength{\tabcolsep}{10pt} 
\begin{tabular}{|l|c|}
\hline
\textbf{Model} & \textbf{Running Time} \\ \hline
Idefics2-8B & 7H 24M \\ 
Intern2-VL-8B & 14H 54M \\ 
Qwen2-VL-Instruct-72B & 5D 3H 50M \\ 
NVLM-D-72B & 5D 0H 13M \\ \hline
\multicolumn{2}{l}{\textbf{Total Questions:} 3.92K} \\
\multicolumn{2}{l}{\textbf{Device:} 2 $\times$ A100 80G GPU} \\ 
\end{tabular}
\caption{Actual Running Time for Each Model. D: Day, H: Hour, M: Minute}
\label{tab:running_time}
\end{table}

\subsection{Discussion Additional Materials}
\subsubsection{Model Parameters Trend}
\label{sec:Model Parameters Trend}
\begin{figure}[ht]
    \centering
    \includegraphics[width=0.45\textwidth]{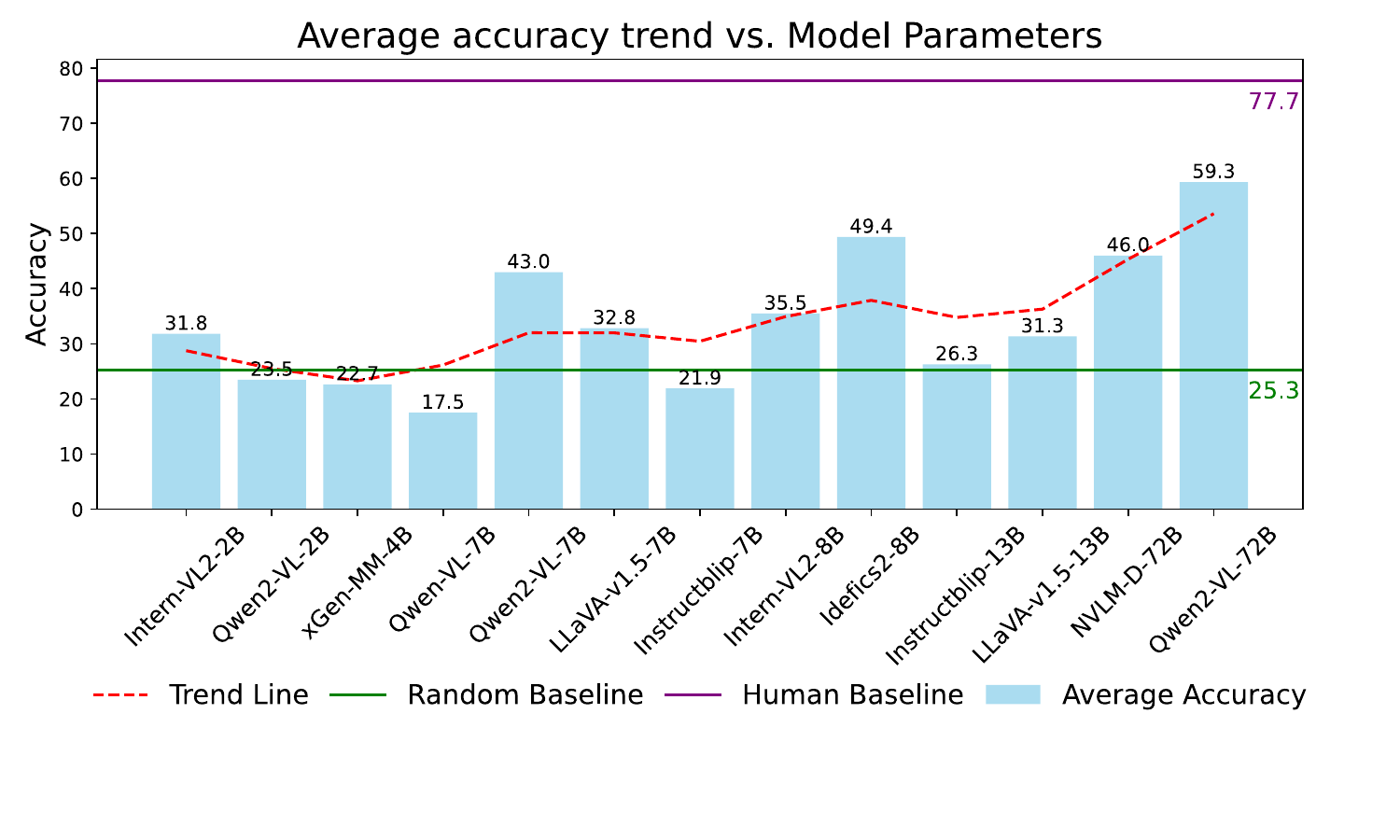}
    \caption{The average accuracy trend for the Direct Answer task as model sizes gradually increase. The trend line is derived using Gaussian smoothing, and the average accuracy is calculated by averaging the results across all five stages.}
    \label{fig:trend_avg}
\end{figure}

\begin{figure}[ht]
    \centering
    \includegraphics[width=0.49\textwidth]{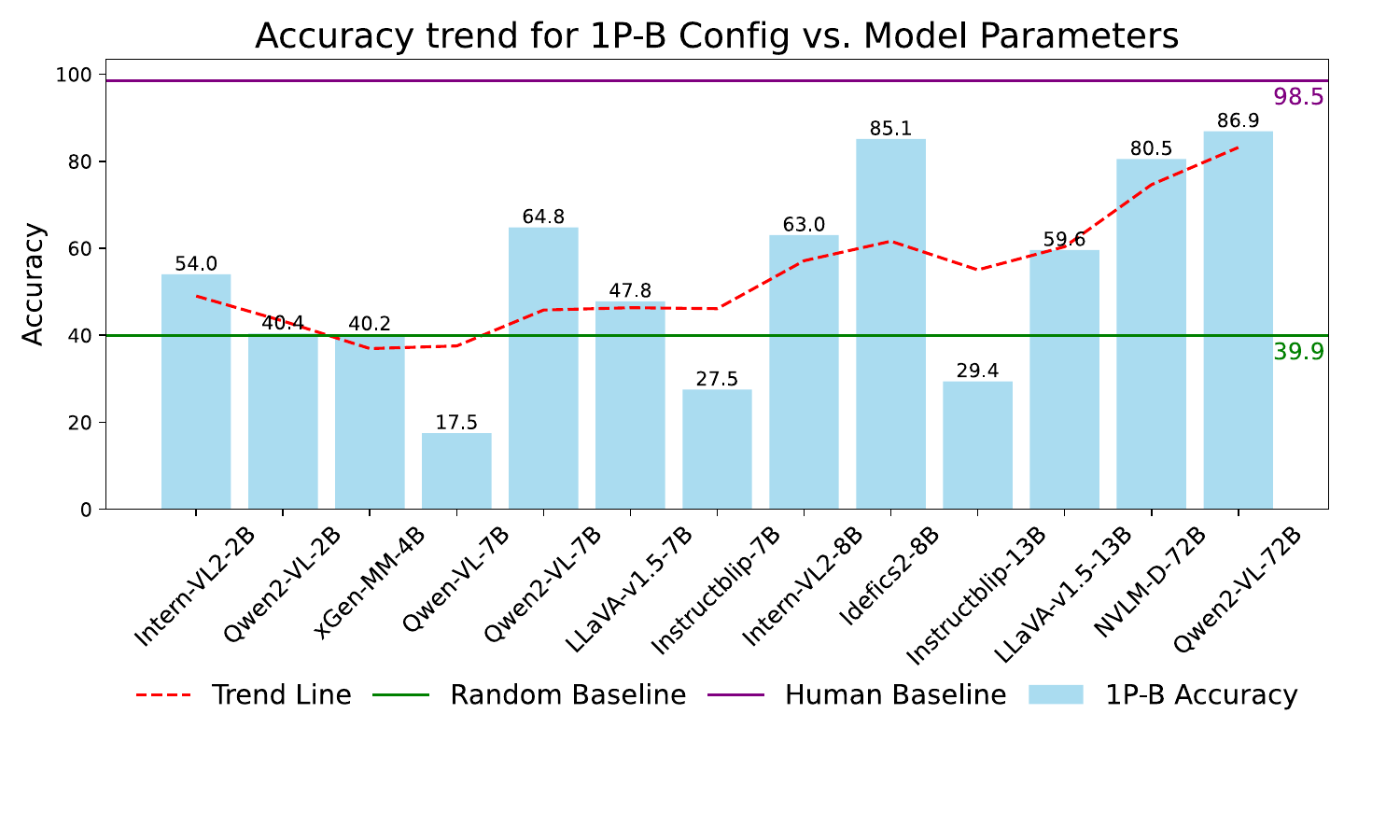}
    \caption{The One-Panel Basic Perception accuracy trend for the Direct Answer task as model sizes gradually increase. The trend line is derived using Gaussian smoothing.}
    \label{fig:trend_1P-B}
\end{figure}

\begin{figure}[ht]
    \centering
    \includegraphics[width=0.49\textwidth]{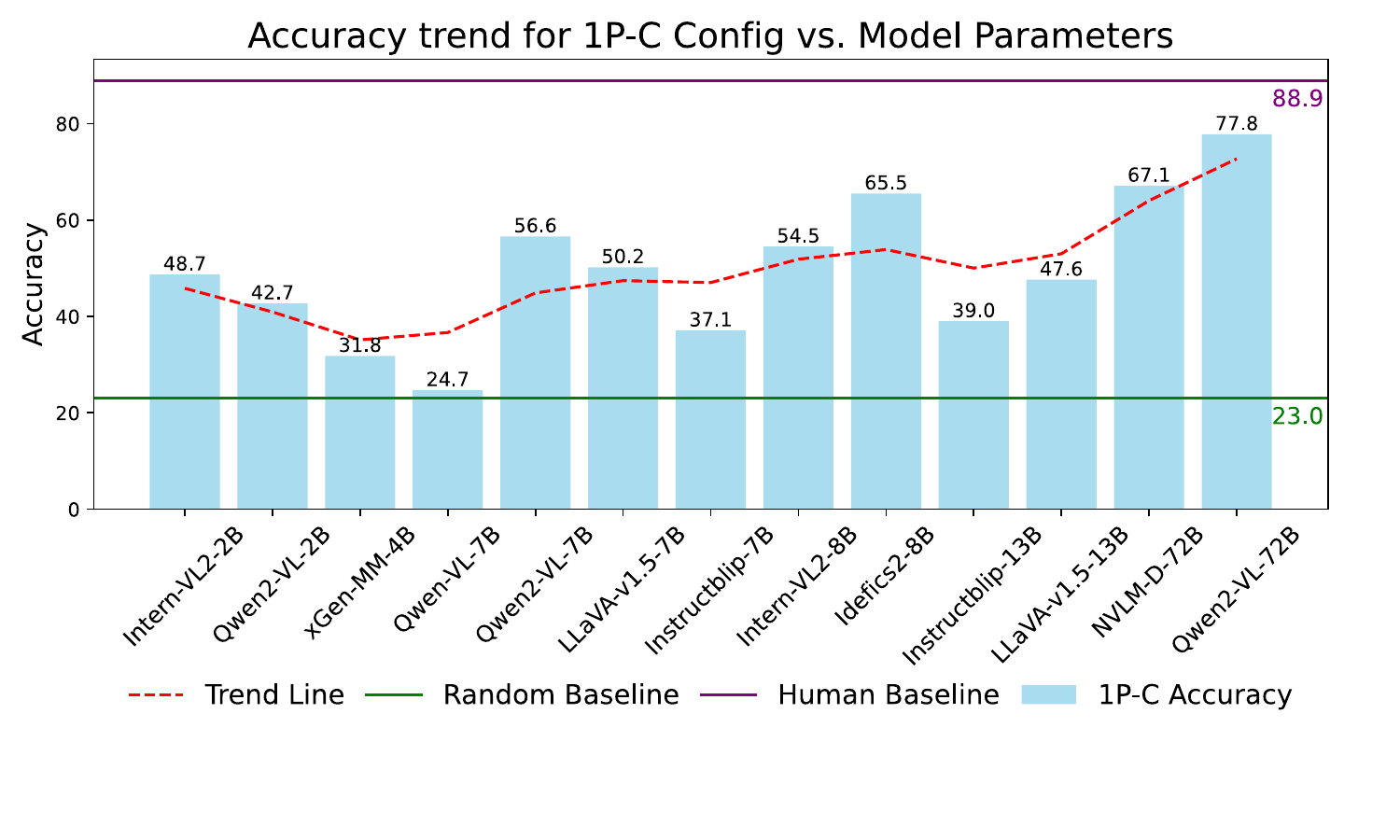}
    \caption{The One-Panel Comparison accuracy trend for the Direct Answer task as model sizes gradually increase. The trend line is derived using Gaussian smoothing.}
    \label{fig:trend_1P-C}
\end{figure}

\begin{figure}[ht]
    \centering
    \includegraphics[width=0.49\textwidth]{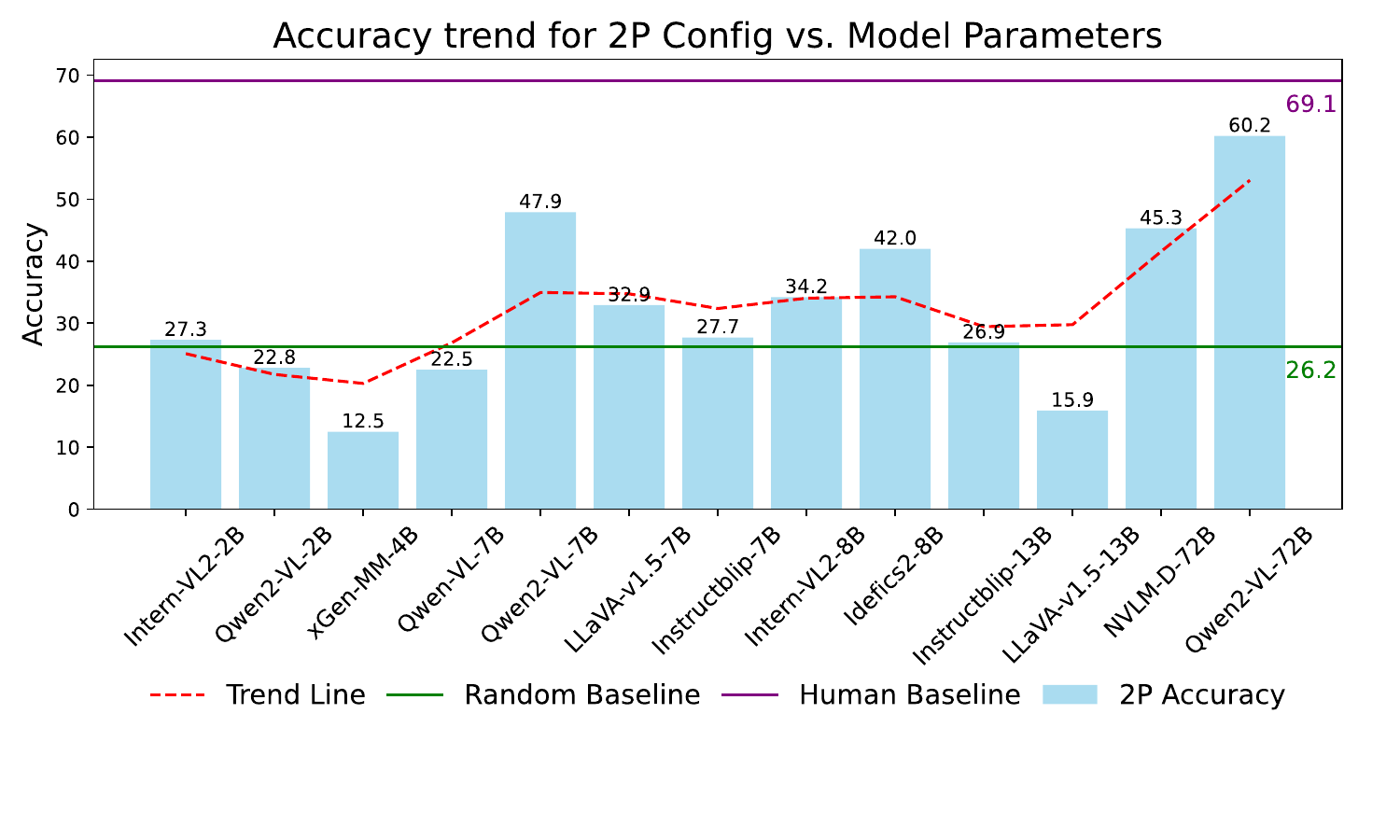}
    \caption{The Two-Panels Comparison accuracy trend for the Direct Answer task as model sizes gradually increase. The trend line is derived using Gaussian smoothing.}
    \label{fig:trend_2P}
\end{figure}

\begin{figure}[ht]
    \centering
    \includegraphics[width=0.49\textwidth]{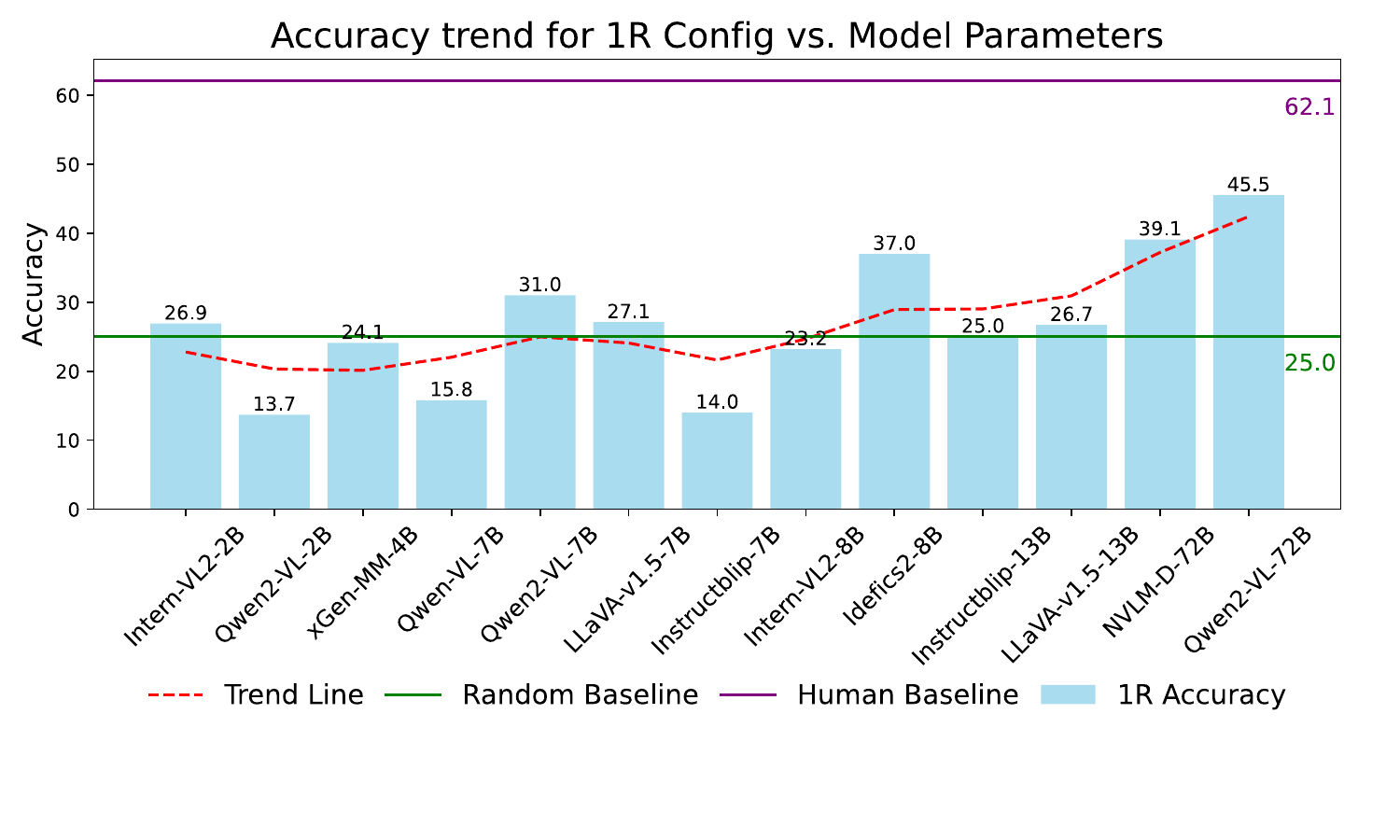}
    \caption{The One-Row Deduction accuracy trend for the Direct Answer task as model sizes gradually increase. The trend line is derived using Gaussian smoothing.}
    \label{fig:trend_1R}
\end{figure}

\begin{figure}[ht]
    \centering
    \includegraphics[width=0.49\textwidth]{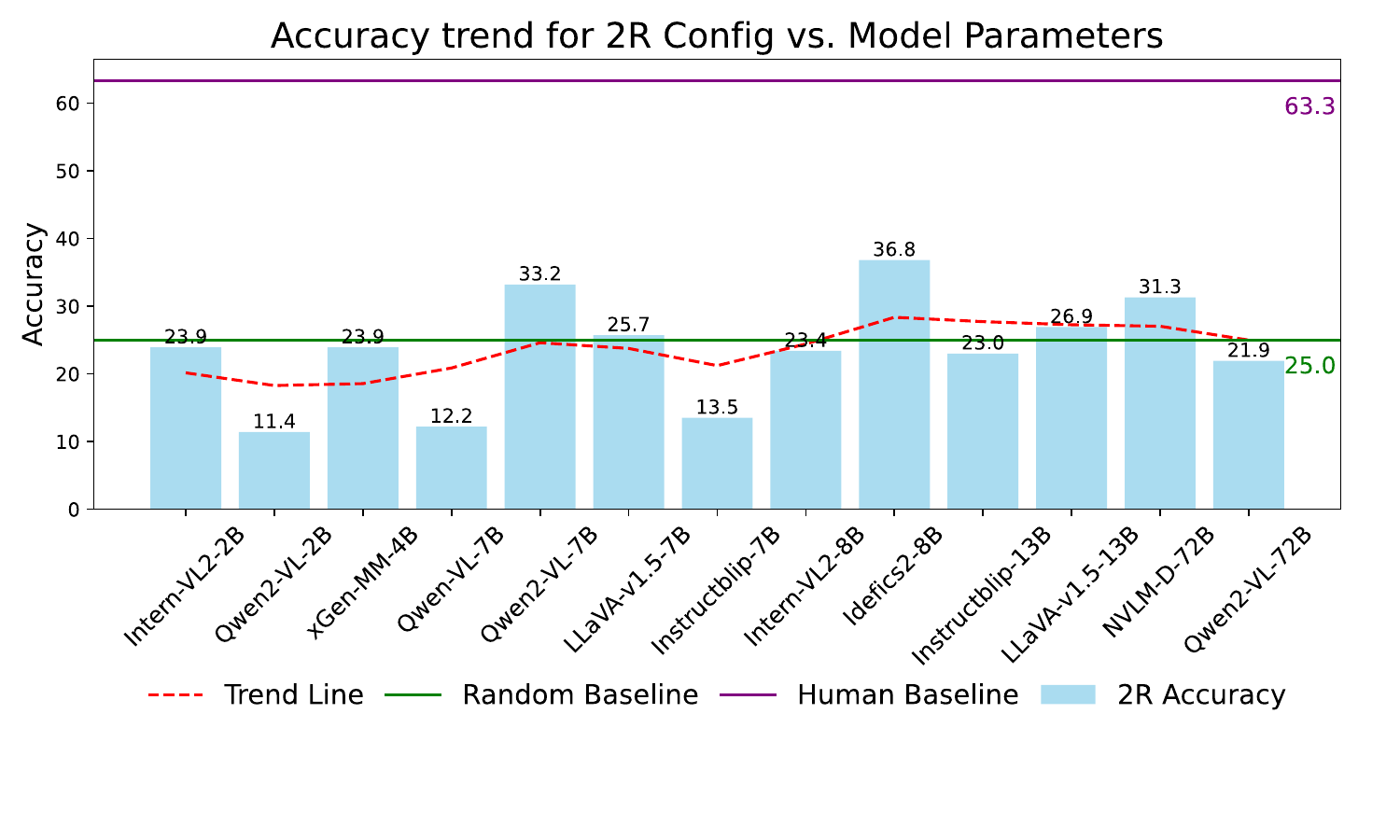}
    \caption{The Two-Rows Deduction accuracy trend for the Direct Answer task as model sizes gradually increase. The trend line is derived using Gaussian smoothing.}
    \label{fig:trend_2R}
\end{figure}

\begin{figure}[ht]
    \centering
    \includegraphics[width=0.49\textwidth]{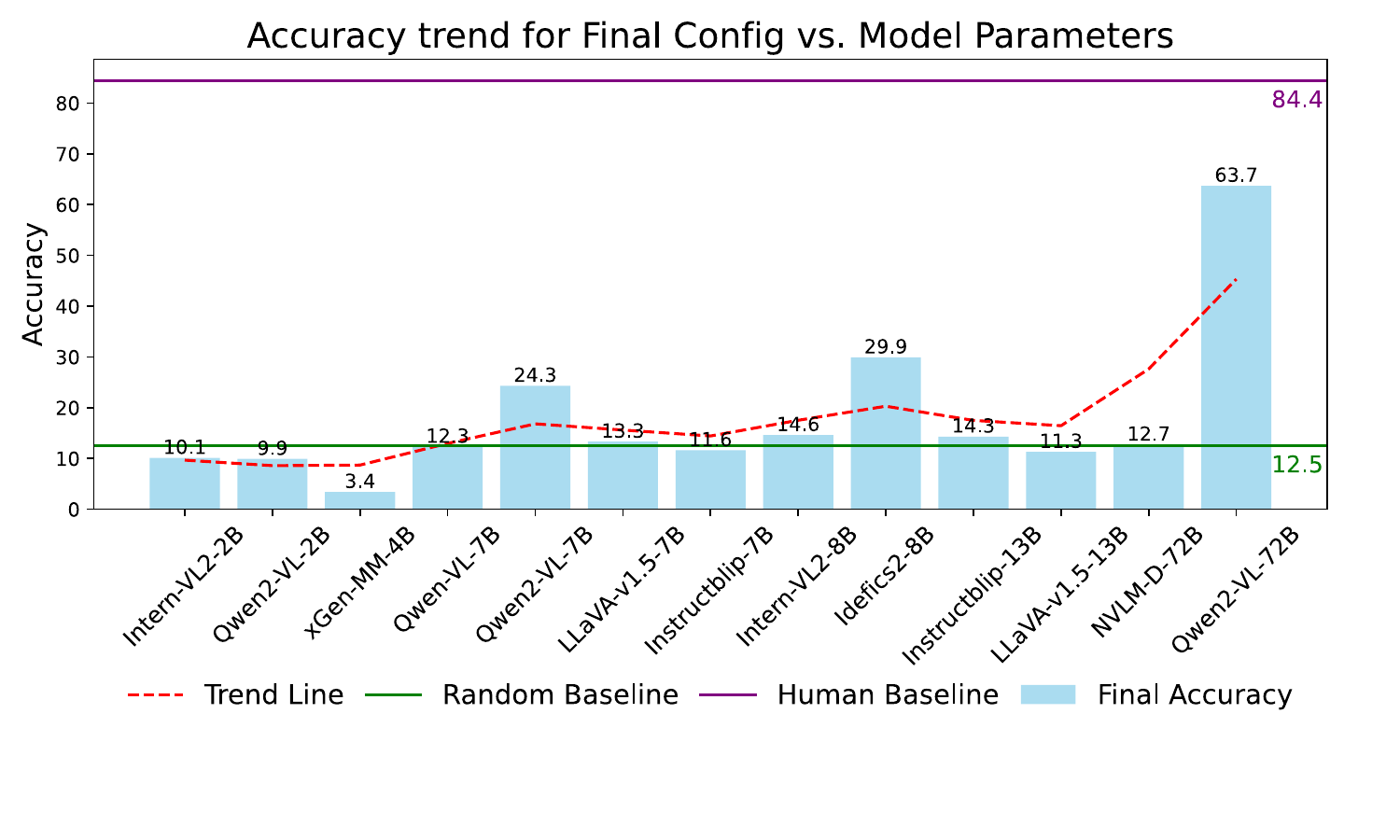}
    \caption{The Final accuracy trend for the Direct Answer task as model sizes gradually increase. The trend line is derived using Gaussian smoothing.}
    \label{fig:trend_Final}
\end{figure}

Figures \ref{fig:trend_avg}, \ref{fig:trend_1P-B}, \ref{fig:trend_1P-C}, \ref{fig:trend_2P}, \ref{fig:trend_1R}, \ref{fig:trend_2R}, and \ref{fig:trend_Final} demonstrate that performance generally improves with larger model parameter sizes across stages, except for the 2R and Final stages. For these two stages, most models perform below the random baseline. The differences in model performance are primarily attributed to the varying sizes of their language encoders, highlighting the significant role of a robust language encoder in overall performance. However, despite the observed improvements, a noticeable gap persists between model and human performance. This discrepancy may arise from the complexity of the visual input, which poses challenges for models in fully understanding and integrating multimodal information.


\subsubsection{Attribute Break-Down Analysis}
\label{sec:Attribute Break-Down Analysis}
\begin{figure*}[ht]
    \centering
    \includegraphics[width=\textwidth]{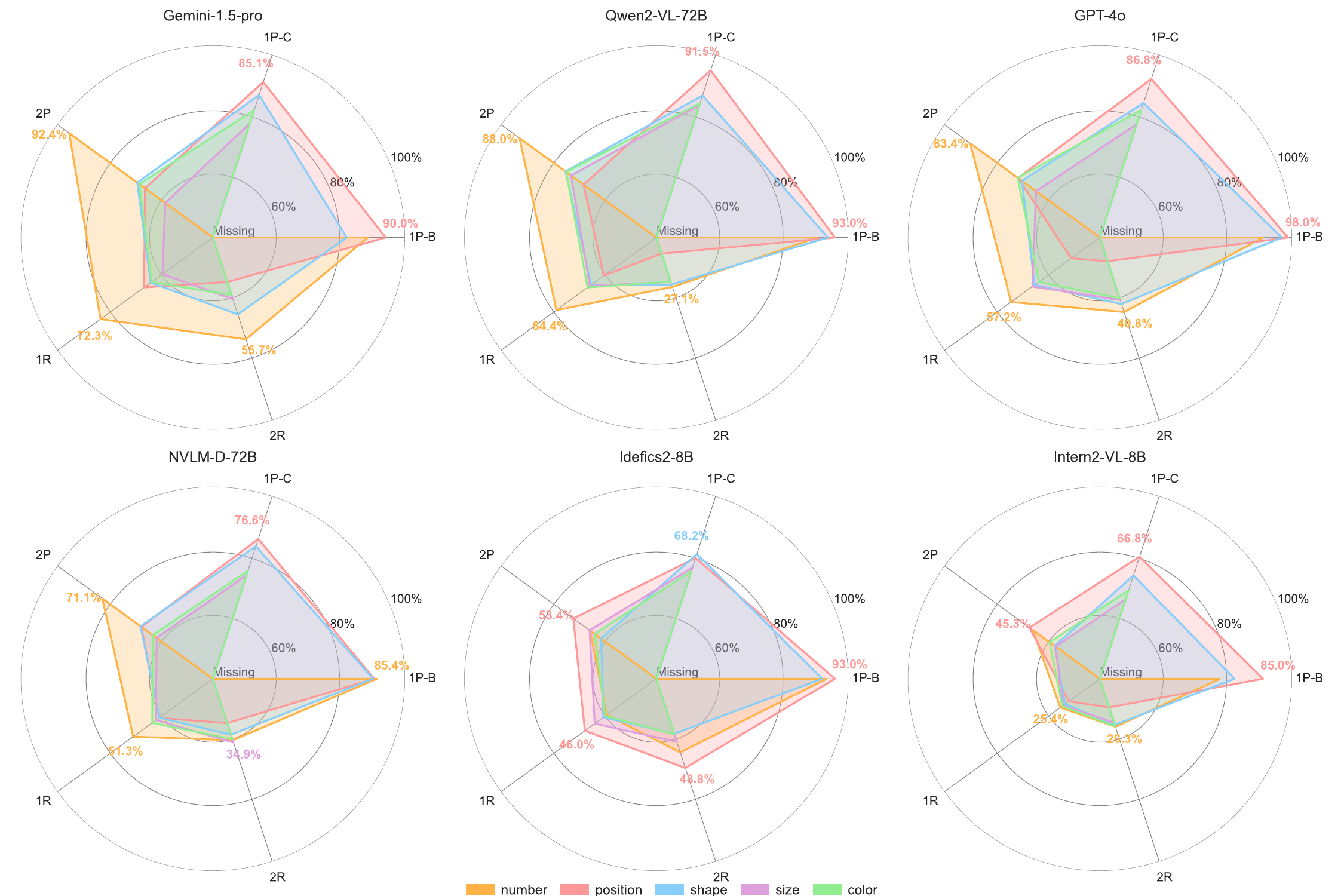}
    \caption{The Radar Chart for all six models in Logical Chain subtask.}
    \label{fig:radar_all}
\end{figure*}

Figure \ref{fig:radar_all} illustrates the attribute-level performance breakdown of two open-source models and four closed-source models evaluated on the logical chain task. Gemini, GPT-4o, and the two larger models, Qwen2-VL and NVLM-D, exhibit similar trends: the Number attribute achieves the highest performance in more complex stages (2P, 1R, and 2R), while Position dominates in lower-level stages (1P-C and 1P-B). In contrast, smaller models like Idefics2 and Intern2-VL struggle with the Number attribute but perform relatively better on Position, indicating that these models are less sensitive to counting tasks but demonstrate better spatial reasoning.

\subsubsection{Handling Long Prompts}
\label{sec:Handling Long Prompts}
\begin{figure}[ht]
    \centering
    \includegraphics[width=\columnwidth]{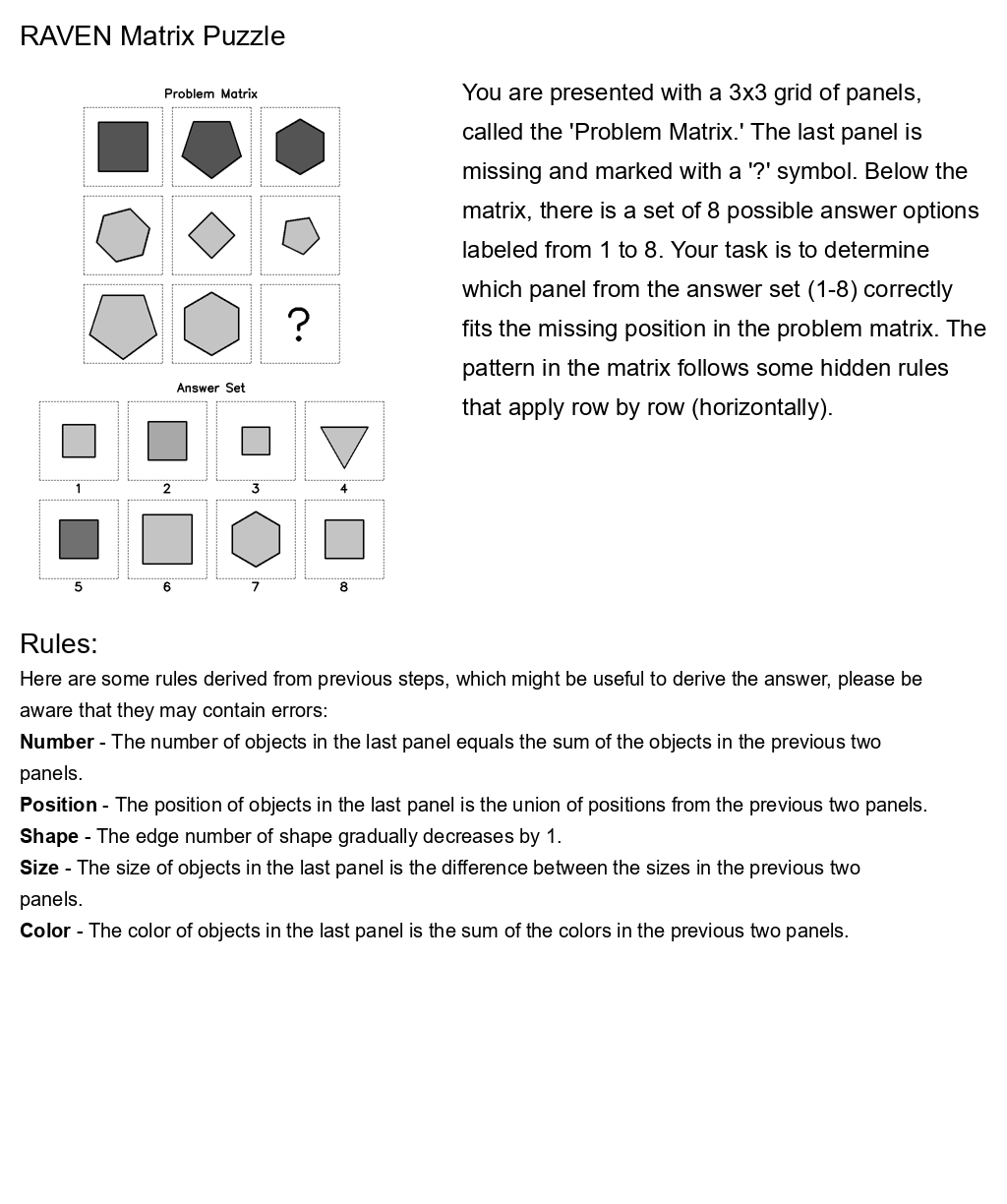}
    \caption{Document Structure for Handling Long Prompt.}
    \label{fig:document}
\end{figure}

Table \ref{table:long prompts extra} presents the accuracy and MSEval scores for three prompting techniques incorporating prior information in the Logical Chain task. For GPT-4o, Qwen2-VL, and Gemini, the use of HTML tags yields significant performance improvements. Additionally, for GPT-4o and Qwen2-VL, the Document-based prompting also demonstrates notable benefits. However, for other models, these two techniques show a negative impact. In this case, the MSEval results are consistent with the accuracy outcomes. List \ref{lst:html} shows an example of HTML structured prompts, while Figure \ref{fig:document} shows an example of Document structured prompts.

\begin{table}[ht]
\setlength{\tabcolsep}{4pt}
\renewcommand{\arraystretch}{1.2}
\resizebox{0.49\textwidth}{!}{
\begin{tabular}{lccccccc}
\hlineB{4}  
                                & Metric          & Prior          & 1P     & 2P     & 1R     & 2R     & Final       \\ \hline\hline
\multirow{3}{*}{GPT-4o}         & \multirow{3}{*}{Acc}   
                                & Vanilla         & 73.8   & 43.9   & 41.8   & 50.6   & 10.0        \\
                                & & Struct.       & \textbf{82.2} & 	64.4 &	47.8 &	50.9 &	8.6           \\
                                & & Doc.        & 80.8 & 	44.8 &	31.1 &	24.9 &	10.0          \\ \cline{1-8}                               
\multirow{3}{*}{Gemini}         & \multirow{3}{*}{Acc}   
                                & Vanilla         & 75.5 & 64.4 & 52.6 & 57.1 & 18.6 \\
                                & & Struct.       & 70.6 &	66.4 &	52.9 &	\textbf{57.8} &	17.1          \\
                                & & Doc.        & 69.6 &	51.0 &	36.7 &	33.1 &	14.3           \\ \cline{1-8}   
\multirow{6}{*}{\makecell{Qwen2-VL \\ (72B)}}   
                                & \multirow{3}{*}{Acc}   
                                & Vanilla         & 74.1 &	57.8 &	47.3 &	54.2 &	65.7$^\ast$ \\
                                & & Struct.       & 77.2 & 	\textbf{67.7} &	\textit{55.1} &	53.6 &	61.4           \\
                                & & Doc.        & 76.5 & 63.1 &	50.2 &	46.6 &	24.3           \\ \cline{2-8}
                                & \multirow{3}{*}{MSEval}     
                                & Vanilla         & 2.54 &	1.95 &	1.79 &	1.70 &	\underline{5.14}$^\ast$ \\
                                & & Struct.       & \underline{2.64} &	\underline{2.46} &	\underline{2.37} &	\underline{2.17} &	3.11           \\
                                & & Doc.        & 2.62 &	2.33 &	2.26 &	1.90 &	1.88           \\ \cline{1-8}
\multirow{6}{*}{\makecell{NVLM-D \\ (72B)}}     
                                & \multirow{3}{*}{Acc}   
                                & Vanilla         & 66.1   & 45.2   & 39.1   & 43.3   & 7.1         \\
                                & & Struct.       & 45.6 & 	25.3 &	23.1 &	36.8 &	10.0           \\
                                & & Doc.        & 18.7 & 11.3 &	17.5 &	14.2 &	20.0          \\ \cline{2-8}
                                & \multirow{3}{*}{MSEval}     
                                & Vanilla         & 2.25 & 	1.20 &	1.28 &	1.02 &	0.76       \\
                                & & Struct.       & 1.84 & 	0.87 &	0.99 &	0.98 &	0.79           \\
                                & & Doc.        & 0.93&  0.78 &	0.88 &	0.94 &	0.96           \\ \cline{1-8}
\multirow{6}{*}{\makecell{Idefics2 \\ (8B)}}   
                                & \multirow{3}{*}{Acc}   
                                & Vanilla         & 57.8 &	37.8 &	36.6 &	42.4 &	25.7 \\
                                & & Struct.       & 46.2 & 	30.5 &	36.9 &	43.5 &	18.6           \\
                                & & Doc.        & 27.2 &	23.6 &	9.6 &	6.9 &	15.7           \\ \cline{2-8}
                                & \multirow{3}{*}{MSEval}     
                                & Vanilla         & 2.02 &	1.48 &	1.51 &	1.51 &	1.44  \\
                                & & Struct.       & 1.59 &	1.18 &	1.25 &	1.33 &	1.27           \\
                                & & Doc.        & 1.04 & 1.01 &	0.97 &	0.98 &	1.00          \\ \cline{1-8}
\multirow{6}{*}{\makecell{Intern2-VL \\ (8B)}}     
                                & \multirow{3}{*}{Acc}   
                                & Vanilla         & 54.4 &	41.9 &	31.6 &	33.5 &	17.1         \\
                                & & Struct.       & 48.4 &	31.0 &	20.7 &	27.3 &	7.1           \\
                                & & Doc.        & 23.1& 30.2 &	17.1 &	17.1 &	8.6           \\ \cline{2-8}
                                & \multirow{3}{*}{MSEval}     
                                & Vanilla         & 2.02 & 	1.48 &	1.51 &	1.51 &	1.44        \\
                                & & Struct.       & 1.52 &	1.18 &	1.10 &	1.00 &	0.97           \\
                                & & Doc.        & 1.02 & 1.00 &	1.00 &	0.97 &	0.92           \\ \hlineB{4}
\end{tabular}
}
\caption{The Accuracy (Acc) and MSEval scores of three prompting techniques for the Logical Chain task. Vanilla:
Pure Text, Struct.: Structure (HTML), Doc.: Document. The highest accuracy are highlighted in \textbf{bold}. The highest MSEval are highlighted in \underline{underline}.}
\label{table:long prompts extra}
\end{table}

\begin{lstlisting}[caption={HTML Structure for Handling Long Prompt}, label={lst:html}]
<!DOCTYPE html>
<html>
  <body>
    <h1>
      In this visual puzzle, you are given two panels. Each panel divided into
      two sections by a vertical line, separating the <strong>left</strong>
      side from the <strong>right</strong> side, with objects might present
      in both sections. Below is the information generated from the previous steps,
      please be aware that it may or may not contain errors:
    </h1>
    <div>
      <h2>Panel Information</h2>
      <ul>
        <li>There are 1 objects in the <strong>left</strong> part of the
        <strong>left</strong> panel.</li>
        <li>There are 2 objects in the <strong>left</strong> part of the
        <strong>right</strong> panel.</li>
      </ul>
    </div>
    <div>
      <h2>Question</h2>
      <p>
        Consider only the <strong>left</strong> part of the two panels in the image.
        Does the <strong>left</strong> panel contain the same number of objects,
        more objects, or fewer objects than the <strong>right</strong> panel?
        Please select one of the following:
      </p>
      <ul>
        <li>A: More</li>
        <li>B: The same</li>
        <li>C: Fewer</li>
      </ul>
      <p>The answer should be one of A, B, C.</p>
    </div>
  </body>
</html>
\end{lstlisting}

\subsection{Qualitative Analysis}
\label{sec:Qualitative Analysis}\

Lists \ref{lst:qualitative_1} to \ref{lst:qualitative_5} present five examples that illustrate the advantages of our MSEval metric over traditional accuracy. For instance, in List \ref{lst:qualitative_4}, all the intermediate steps are incorrect, yet the model arrives at the correct answer with only a small confidence margin (31.375 compared to the second-highest confidence of 31.125). In this case, the model's performance is not truly effective, as it is unclear how it managed to produce the correct answer despite incorrect intermediate steps. Unlike traditional accuracy, which would mark this as fully correct, MSEval appropriately penalizes such cases by assigning a low score.

Conversely, a reverse scenario is shown in List \ref{lst:qualitative_3}, where traditional accuracy marks the result as entirely incorrect. However, the model correctly answers all intermediate steps, and the probability of the correct answer is very close to the highest confidence value. In this situation, MSEval assigns a relatively high score, reflecting the model's partial success and rewarding its correct reasoning process.

\clearpage
\onecolumn
\input{prompt}

\end{document}

%% file: prompt.tex
\lstset{
    basicstyle=\fontsize{10}{10}\selectfont\ttfamily, 
    frame=single,                                     
    breaklines=true,                                  
    breakindent=0ex,                                  
    xleftmargin=8pt,                                 
    xrightmargin=8pt,                                
    backgroundcolor=\color[gray]{0.95},               
    keywordstyle=\color{blue}\bfseries,              
    stringstyle=\color{red},                          
    commentstyle=\color{gray},                        
    columns=fullflexible                             
}
\definecolor{mydarkgreen}{RGB}{0, 100, 0}
\definecolor{mydarkred}{RGB}{139, 0, 0}
\lstset{escapeinside={<@}{@>}}
\begin{lstlisting}[
    caption={An instance of high accuracy but low MSEval occurs since the LLM NVLM-D-72B generates a current-stage answer consistent with the ground truth, while earlier dependent stages produce inconsistent results.},
    captionpos=b, label={lst:qualitative_1}
]
---------------------------------
Dependent Stage Name:  single_panel_1_left
Dependent Stage Question:  Are all objects in the left part of the panel of the same color?
Dependent Stage Choice:  ['A: Only one object', 'B: No', 'C: Yes']
<@\textbf{\textcolor{mydarkgreen}{Dependent Stage Ground Truth:  A}}@>
Dependent Stage Logits:  {'A': 11.125, 'B': 11.125, 'C': 20.75}
<@\textbf{\textcolor{mydarkred}{Dependent Stage Generated Answer:  C}}@>
---------------------------------
Dependent Stage Name:  single_panel_2_left
Dependent Stage Question:  Are all objects in the left part of the panel of the same color?
Dependent Stage Choice:  ['A: No', 'B: Only one object', 'C: Yes']
<@\textbf{\textcolor{mydarkgreen}{Dependent Stage Ground Truth:  B}}@>
Dependent Stage Logits:  {'A': 11.125, 'B': 10.5625, 'C': 19.875}
<@\textbf{\textcolor{mydarkred}{Dependent Stage Generated Answer: C}}@>
---------------------------------
Dependent Stage Name:  single_panel_3_left
Dependent Stage Question:  Are all objects in the left part of the panel of the same color?
Dependent Stage Choice:  ['A: Only one object', 'B: No', 'C: Yes']
<@\textbf{\textcolor{mydarkgreen}{Dependent Stage Ground Truth:  A}}@>
Dependent Stage Logits:  {'A': 10.375, 'B': 10.375, 'C': 19.75}
<@\textbf{\textcolor{mydarkred}{Dependent Stage Generated Answer:  C}}@>
---------------------------------
Dependent Stage Name:  two_panels_1_2_left
Dependent Stage Question:  Consider only the left part of the two panels in the image. Is the color of all the objects in the left panel the same as, darker or brighter than the objects in the right panel? If the colors within either panel are already different from each other, select 'Not Comparable.'
Dependent Stage Choice:  ['A: Not comparable', 'B: The same', 'C: Darker', 'D: Brighter']
<@\textbf{\textcolor{mydarkgreen}{Dependent Stage Ground Truth:  C}}@>
Dependent Stage Logits:  {'A': 16.5, 'B': 17.75, 'C': 16.375, 'D': 15.0625}
<@\textbf{\textcolor{mydarkred}{Dependent Stage Generated Answer:  B}}@>
---------------------------------
Dependent Stage Name:  two_panels_2_3_left
Dependent Stage Question:  Consider only the left part of the two panels in the image. Is the color of all the objects in the left panel the same as, darker or brighter than the objects in the right panel? If the colors within either panel are already different from each other, select 'Not Comparable.'
Dependent Stage Choice:  ['A: Not comparable', 'B: Darker', 'C: Brighter', 'D: The same']
<@\textbf{\textcolor{mydarkgreen}{Dependent Stage Ground Truth:  B}}@>
Dependent Stage Logits:  {'A': 18.875, 'B': 18.0, 'C': 16.125, 'D': 19.5}
<@\textbf{\textcolor{mydarkred}{Dependent Stage Generated Answer:  D}}@>
---------------------------------
Current Stage:
Current Stage Name:  one_row_left
Current Stage Question:  Look at the three panels in the image from left to right, paying attention only to the left portions of each panel, and identify the rule that controls the color of objects.
Current Stage Choice:  ['A: The color of objects in the last panel is the sum of the colors in the previous two panels.', 'B: The color of objects gradually brightens by a constant amount each time.', 'C: The color of objects gradually darkens by a constant amount each time.', 'D: The color of objects in the last panel is the difference between the colors in the previous two panels.']
<@\textbf{\textcolor{mydarkgreen}{Current Stage Ground Truth:  B}}@>
Current Stage Logits:  {'A': 20.0, 'B': 21.375, 'C': 20.5, 'D': 19.75}
<@\textbf{\textcolor{mydarkgreen}{Current Stage Generated Answer:  B}}@>
---------------------------------
Accuracy:  1.0
MSEval:  1.066
MSEval Random Baseline:  1.0
\end{lstlisting}

\clearpage

\begin{lstlisting}[
    caption={An instance of low accuracy but high MSEval arises as the LLM NVLM-D-72B generates a current-stage answer inconsistent with the ground truth, despite earlier dependent stages producing consistent results.},
    captionpos=b, label={lst:qualitative_2} 
]
---------------------------------
Dependent Stage Name:  single_panel_1_left
Dependent Stage Question:  What is the shape of the object in the left part of the panel?
Dependent Stage Choice:  ['A: circle', 'B: hexagon', 'C: triangle', 'D: square']
<@\textbf{\textcolor{mydarkgreen}{Dependent Stage Ground Truth:  B}}@>
Dependent Stage Logits:  {'A': 17.125, 'B': 24.5, 'C': 16.0, 'D': 16.0}
<@\textbf{\textcolor{mydarkgreen}{Dependent Stage Generated Answer:  B}}@>
---------------------------------
Dependent Stage Name:  single_panel_2_left
Dependent Stage Question:  What is the shape of the object in the left part of the panel?
Dependent Stage Choice:  ['A: triangle', 'B: pentagon', 'C: square', 'D: hexagon']
<@\textbf{\textcolor{mydarkgreen}{Dependent Stage Ground Truth:  B}}@>
Dependent Stage Logits:  {'A': 17.5, 'B': 23.875, 'C': 17.125, 'D': 15.8125}
<@\textbf{\textcolor{mydarkgreen}{Dependent Stage Generated Answer:  B}}@>
---------------------------------
Dependent Stage Name:  single_panel_3_left
Dependent Stage Question:  What is the shape of the object in the left part of the panel?
Dependent Stage Choice:  ['A: hexagon', 'B: pentagon', 'C: square', 'D: triangle']
<@\textbf{\textcolor{mydarkgreen}{Dependent Stage Ground Truth:  C}}@>
Dependent Stage Logits:  {'A': 17.375, 'B': 16.875, 'C': 24.125, 'D': 17.375}
<@\textbf{\textcolor{mydarkgreen}{Dependent Stage Generated Answer:  C}}@>
---------------------------------
Dependent Stage Name:  two_panels_1_2_left
Dependent Stage Question:  Consider only the left part of the two panels in the image. Is the shape of all the objects in the left panel have the same, more, or fewer edges compared to the objects in the right panel? If the shapes within either panel are already different from each other, select 'Not Comparable.' (Note: The edge number increases in the following order: triangle, square, pentagon, hexagon, circle)
Dependent Stage Choice:  ['A: Not comparable', 'B: Fewer', 'C: The same', 'D: More']
<@\textbf{\textcolor{mydarkgreen}{Dependent Stage Ground Truth:  D}}@>
Dependent Stage Logits:  {'A': 21.625, 'B': 20.875, 'C': 20.0, 'D': 22.25}
<@\textbf{\textcolor{mydarkgreen}{Dependent Stage Generated Answer:  D}}@>
---------------------------------
Dependent Stage Name:  two_panels_2_3_left
Dependent Stage Question:  Consider only the left part of the two panels in the image. Is the shape of all the objects in the left panel have the same, more, or fewer edges compared to the objects in the right panel? If the shapes within either panel are already different from each other, select 'Not Comparable.' (Note: The edge number increases in the following order: triangle, square, pentagon, hexagon, circle)
Dependent Stage Choice:  ['A: The same', 'B: Not comparable', 'C: More', 'D: Fewer']
<@\textbf{\textcolor{mydarkgreen}{Dependent Stage Ground Truth:  C}}@>
Dependent Stage Logits:  {'A': 20.875, 'B': 21.75, 'C': 21.875, 'D': 20.25}
<@\textbf{\textcolor{mydarkgreen}{Dependent Stage Generated Answer:  C}}@>
---------------------------------
Current Stage:
Current Stage Name:  one_row_left
Current Stage Question:  Look at the three panels in the image from left to right, paying attention only to the left portions of each panel, and identify the rule that controls the shape of objects. (Note: The edge number increases in the following order: triangle, square, pentagon, hexagon, circle)
Current Stage Choice:  ['A: The edge number of shape gradually decreases by 1.', 'B: The edge number of shape gradually increases by 1.', 'C: No clear rule is present.', 'D: The shape remains constant.']
<@\textbf{\textcolor{mydarkgreen}{Current Stage Ground Truth:  A}}@>
Current Stage Logits:  {'A': 21.75, 'B': 22.75, 'C': 19.25, 'D': 16.625}
<@\textbf{\textcolor{mydarkred}{Current Stage Generated Answer:  B}}@>
---------------------------------
Accuracy:  0.0
MSEval:  2.506839853582758
MSEval Random Baseline:  1.0
\end{lstlisting}
\clearpage
\begin{lstlisting}[
caption={An instance of low accuracy but high MSEval arises as the LLM Intern-VL2-8B generates a current-stage answer inconsistent with the ground truth, despite earlier dependent stages producing consistent results.},
captionpos=b, label={lst:qualitative_3}
]
---------------------------------
Dependent Stage Name:  single_panel_1_left
Dependent Stage Question:  What is the shape of the object in the left part of the panel?
Dependent Stage Choice:  ['A: circle', 'B: hexagon', 'C: triangle', 'D: square']
<@\textbf{\textcolor{mydarkgreen}{Dependent Stage Ground Truth:  B}}@>
Dependent Stage Logits:  {'A': 21.125, 'B': 26.375, 'C': 20.375, 'D': 22.375}
<@\textbf{\textcolor{mydarkgreen}{Dependent Stage Generated Answer:  B}}@>
---------------------------------
Dependent Stage Name:  single_panel_2_left
Dependent Stage Question:  What is the shape of the object in the left part of the panel?
Dependent Stage Choice:  ['A: triangle', 'B: pentagon', 'C: square', 'D: hexagon']
<@\textbf{\textcolor{mydarkgreen}{Dependent Stage Ground Truth:  B}}@>
Dependent Stage Logits:  {'A': 22.5, 'B': 25.5, 'C': 21.25, 'D': 24.625}
<@\textbf{\textcolor{mydarkgreen}{Dependent Stage Generated Answer:  B}}@>
---------------------------------
Dependent Stage Name:  single_panel_3_left
Dependent Stage Question:  What is the shape of the object in the left part of the panel?
Dependent Stage Choice:  ['A: hexagon', 'B: pentagon', 'C: square', 'D: triangle']
<@\textbf{\textcolor{mydarkgreen}{Dependent Stage Ground Truth:  C}}@>
Dependent Stage Logits:  {'A': 22.875, 'B': 21.125, 'C': 27.5, 'D': 23.625}
<@\textbf{\textcolor{mydarkgreen}{Dependent Stage Generated Answer:  C}}@>
---------------------------------
Dependent Stage Name:  two_panels_1_2_left
Dependent Stage Question:  Consider only the left part of the two panels in the image. Is the shape of all the objects in the left panel have the same, more, or fewer edges compared to the objects in the right panel? If the shapes within either panel are already different from each other, select 'Not Comparable.' (Note: The edge number increases in the following order: triangle, square, pentagon, hexagon, circle)
Dependent Stage Choice:  ['A: Not comparable', 'B: Fewer', 'C: The same', 'D: More']
<@\textbf{\textcolor{mydarkgreen}{Dependent Stage Ground Truth:  D}}@>
Dependent Stage Logits:  {'A': 30.5, 'B': 30.25, 'C': 30.75, 'D': 30.875}
<@\textbf{\textcolor{mydarkgreen}{Dependent Stage Generated Answer:  D}}@>
---------------------------------
Dependent Stage Name:  two_panels_2_3_left
Dependent Stage Question:  Consider only the left part of the two panels in the image. Is the shape of all the objects in the left panel have the same, more, or fewer edges compared to the objects in the right panel? If the shapes within either panel are already different from each other, select 'Not Comparable.' (Note: The edge number increases in the following order: triangle, square, pentagon, hexagon, circle)
Dependent Stage Choice:  ['A: The same', 'B: Not comparable', 'C: More', 'D: Fewer']
<@\textbf{\textcolor{mydarkgreen}{Dependent Stage Ground Truth:  C}}@>
Dependent Stage Logits:  {'A': 30.375, 'B': 29.75, 'C': 30.375, 'D': 30.25}
<@\textbf{\textcolor{mydarkgreen}{Dependent Stage Generated Answer:  C}}@>
---------------------------------
Current Stage:
Current Stage Name:  one_row_left
Current Stage Question:  Look at the three panels in the image from left to right, paying attention only to the left portions of each panel, and identify the rule that controls the shape of objects. (Note: The edge number increases in the following order: triangle, square, pentagon, hexagon, circle)
Current Stage Choice:  ['A: The edge number of shape gradually decreases by 1.', 'B: The edge number of shape gradually increases by 1.', 'C: No clear rule is present.', 'D: The shape remains constant.']
<@\textbf{\textcolor{mydarkgreen}{Current Stage Ground Truth:  A}}@>
Current Stage Logits:  {'A': 32.75, 'B': 33.0, 'C': 31.25, 'D': 30.5}
<@\textbf{\textcolor{mydarkred}{Current Stage Generated Answer:  B}}@>
---------------------------------
Accuracy:  0.0
MSEval:  2.228
MSEval Random Baseline: 1.0
\end{lstlisting}

\clearpage
\begin{lstlisting}[
caption={An instance of high accuracy but low MSEval occurs since the LLM Intern-VL2-8B generates a current-stage answer consistent with the ground truth, while earlier dependent stages produce inconsistent results.},
captionpos=b, label={lst:qualitative_4}
]
---------------------------------
Dependent Stage Name:  single_panel_1_right
Dependent Stage Question:  Are all objects in the right part of the panel of the same size?
Dependent Stage Choice:  ['A: No', 'B: Yes', 'C: Only one object']
<@\textbf{\textcolor{mydarkgreen}{Dependent Stage Ground Truth:  C}}@>
Dependent Stage Logits:  {'A': 31.125, 'B': 31.125, 'C': 29.625}
<@\textbf{\textcolor{mydarkred}{Dependent Stage Generated Answer:  A}}@>
---------------------------------
Dependent Stage Name:  single_panel_2_right
Dependent Stage Question:  Are all objects in the right part of the panel of the same size?
Dependent Stage Choice:  ['A: No', 'B: Yes', 'C: Only one object']
<@\textbf{\textcolor{mydarkgreen}{Dependent Stage Ground Truth:  C}}@>
Dependent Stage Logits:  {'A': 31.0, 'B': 31.125, 'C': 29.5}
<@\textbf{\textcolor{mydarkred}{Dependent Stage Generated Answer:  B}}@>
---------------------------------
Dependent Stage Name:  single_panel_3_right
Dependent Stage Question:  Are all objects in the right part of the panel of the same size?
Dependent Stage Choice:  ['A: Only one object', 'B: No', 'C: Yes']
<@\textbf{\textcolor{mydarkgreen}{Dependent Stage Ground Truth:  A}}@>
Dependent Stage Logits:  {'A': 31.375, 'B': 31.375, 'C': 31.625}
<@\textbf{\textcolor{mydarkred}{Dependent Stage Generated Answer:  C}}@>
---------------------------------
Dependent Stage Name:  two_panels_1_2_right
Dependent Stage Question:  Consider only the right part of the two panels in the image. Is the size of all the objects in the left panel the same as, smaller or larger than the objects in the right panel? If the sizes within either panel are already different from each other, select 'Not Comparable.
Dependent Stage Choice:  ['A: Not comparable', 'B: Smaller', 'C: Larger', 'D: The same']
<@\textbf{\textcolor{mydarkgreen}{Dependent Stage Ground Truth:  C}}@>
Dependent Stage Logits:  {'A': 30.875, 'B': 30.75, 'C': 30.25, 'D': 30.75}
<@\textbf{\textcolor{mydarkred}{Dependent Stage Generated Answer:  A}}@>
---------------------------------
Dependent Stage Name:  two_panels_2_3_right
Dependent Stage Question:  Consider only the right part of the two panels in the image. Is the size of all the objects in the left panel the same as, smaller or larger than the objects in the right panel? If the sizes within either panel are already different from each other, select 'Not Comparable.
Dependent Stage Choice:  ['A: Not comparable', 'B: Smaller', 'C: The same', 'D: Larger']
<@\textbf{\textcolor{mydarkgreen}{Dependent Stage Ground Truth:  D}}@>
Dependent Stage Logits:  {'A': 30.25, 'B': 30.125, 'C': 30.5, 'D': 30.125}
<@\textbf{\textcolor{mydarkred}{Dependent Stage Generated Answer:  C}}@>
---------------------------------
Current Stage:
Current Stage Name:  one_row_right
Current Stage Question:  Analyze the three panels in the image from left to right, concentrating only on the right areas of each panel, and determine the rule that dictates the size of objects.
Current Stage Choice:  ['A: The size of objects gradually decreases by a constant amount each time.', 'B: The size of objects in the last panel is the difference between the sizes in the previous two panels.', 'C: The size remains constant.', 'D: Three distinct sizes across panels, rotating through each possible permutation of these sizes.']
<@\textbf{\textcolor{mydarkgreen}{Current Stage Ground Truth:  D}}@>
Current Stage Logits:  {'A': 31.125, 'B': 30.375, 'C': 31.0, 'D': 31.375}
<@\textbf{\textcolor{mydarkgreen}{Current Stage Generated Answer:  D}}@>
---------------------------------
Accuracy:  1.0
MSEval:  0.955
MSEval Random Baseline: 1.0
\end{lstlisting}
\clearpage

\begin{lstlisting}[
    caption={An instance of low accuracy but high MSEval arises as the LLM Qwen2-VL-72B generates a current-stage answer inconsistent with the ground truth, despite earlier dependent stages producing consistent results.},
    captionpos=b, label={lst:qualitative_5}
]
---------------------------------
Dependent Stage Name:  single_panel_1_right
Dependent Stage Question:  What is the shape of the object in the right part of the panel?
Dependent Stage Choice:  ['A: circle', 'B: square', 'C: hexagon', 'D: pentagon']
<@\textbf{\textcolor{mydarkgreen}{Dependent Stage Ground Truth:  C}}@>
Dependent Stage Logits:  {'A': 0, 'B': 0, 'C': 30.625, 'D': 0}
<@\textbf{\textcolor{mydarkgreen}{Dependent Stage Generated Answer:  C}}@>
---------------------------------
Dependent Stage Name:  single_panel_2_right
Dependent Stage Question:  What is the shape of the object in the right part of the panel?
Dependent Stage Choice:  ['A: pentagon', 'B: triangle', 'C: square', 'D: circle']
<@\textbf{\textcolor{mydarkgreen}{Dependent Stage Ground Truth:  B}}@>
Dependent Stage Logits:  {'A': 0, 'B': 30.5, 'C': 0, 'D': 0}
<@\textbf{\textcolor{mydarkgreen}{Dependent Stage Generated Answer:  B}}@>
---------------------------------
Dependent Stage Name:  single_panel_3_right
Dependent Stage Question:  What is the shape of the object in the right part of the panel?
Dependent Stage Choice:  ['A: triangle', 'B: hexagon', 'C: square', 'D: circle']
<@\textbf{\textcolor{mydarkgreen}{Dependent Stage Ground Truth:  C}}@>
Dependent Stage Logits:  {'A': 0, 'B': 0, 'C': 30.875, 'D': 0}
<@\textbf{\textcolor{mydarkgreen}{Dependent Stage Generated Answer:  C}}@>
---------------------------------
Dependent Stage Name:  two_panels_1_2_right
Dependent Stage Question:  Consider only the right part of the two panels in the image. Is the shape of all the objects in the left panel have the same, more, or fewer edges compared to the objects in the right panel? If the shapes within either panel are already different from each other, select 'Not Comparable.' (Note: The edge number increases in the following order: triangle, square, pentagon, hexagon, circle)
Dependent Stage Choice:  ['A: The same', 'B: More', 'C: Fewer', 'D: Not comparable']
<@\textbf{\textcolor{mydarkgreen}{Dependent Stage Ground Truth:  B}}@>
Dependent Stage Logits:  {'A': 0, 'B': 30.0, 'C': 0, 'D': 0}
<@\textbf{\textcolor{mydarkgreen}{Dependent Stage Generated Answer:  B}}@>
---------------------------------
Dependent Stage Name:  two_panels_2_3_right
Dependent Stage Question:  Consider only the right part of the two panels in the image. Is the shape of all the objects in the left panel have the same, more, or fewer edges compared to the objects in the right panel? If the shapes within either panel are already different from each other, select 'Not Comparable.' (Note: The edge number increases in the following order: triangle, square, pentagon, hexagon, circle)
Dependent Stage Choice:  ['A: Not comparable', 'B: Fewer', 'C: More', 'D: The same']
<@\textbf{\textcolor{mydarkgreen}{Dependent Stage Ground Truth:  B}}@>
Dependent Stage Logits:  {'A': 0, 'B': 28.25, 'C': 0, 'D': 0}
<@\textbf{\textcolor{mydarkgreen}{Dependent Stage Generated Answer:  B}}@>
---------------------------------
Current Stage:
Current Stage Name:  one_row_right
Current Stage Question:  Inspect the three panels in the image from left to right, focusing exclusively on the right parts of each panel, and uncover the rule that governs the shape of objects. (Note: The edge number increases in the following order: triangle, square, pentagon, hexagon, circle)
Current Stage Choice:  ['A: The shape remains constant.', 'B: Three distinct shapes across panels, rotating through each possible permutation of these shapes.', 'C: No clear rule is present.', 'D: The edge number of shape gradually increases by 1.']
<@\textbf{\textcolor{mydarkgreen}{Current Stage Ground Truth:  B}}@>
Current Stage Logits:  {'A': 0, 'B': 0, 'C': 31.67, 'D': 0}
<@\textbf{\textcolor{mydarkred}{Current Stage Generated Answer:  C}}@>
---------------------------------
Accuracy:  0.0
MSEval:  2.345087186311839
MSEval Random Baseline:  1.0
\end{lstlisting}